\documentclass[default,iicol]{sn-jnl}% Default with double column layout

\usepackage{mathtools}
\usepackage{caption}
\usepackage{subcaption}

\usepackage{color}
\usepackage{etoolbox}

\definecolor{MyraFGcolor}{RGB}{0,140,0} % darkgreen
\definecolor{ClaraFGcolor}{RGB}{0,100,240} % mediumblue

\definecolor{ForMyraBGcolor}{RGB}{100,150,0} % palegreen
\definecolor{ForClaraBGcolor}{RGB}{175,238,238} % paleturquoise; feel free to change

\definecolor{anyFGcolor}{RGB}{255,100,0}  % orange
\definecolor{anyBGcolor}{RGB}{240,240,00}

\newcommand{\cobalt}{Cost-Based Layer Selector}
\newcommand{\COBALT}{COBALT}

\DeclarePairedDelimiter\abs{\lvert}{\rvert}%
\jyear{2021}%

\theoremstyle{thmstyleone}%
\theoremstyle{thmstyletwo}%

\theoremstyle{thmstylethree}%

\begin{document}

\title[Discovery of patient phenotypes]{A cost-based multi-layer network approach for the discovery of patient phenotypes}


\author*[1]{\fnm{Clara} \sur{Puga}}\email{clara.puga@ovgu.de}

\author[1]{\fnm{Uli} \sur{Niemann}}\email{uli.niemann@ovgu.de}

\author[2]{\fnm{Winfried} \sur{Schlee}}\email{winfried.schlee@gmail.com}

\author[1]{\fnm{Myra} \sur{Spiliopoulou}}\email{myra@ovgu.de}

\affil[1]{\orgdiv{Knowledge Management \&  Discovery Lab}, \orgname{Otto-von-Guericke University}, \orgaddress{\street{Gustav-Adolf-Strasse 15}, \city{Magdeburg}, \postcode{39106}, \state{Saxony-Anhalt}, \country{Germany}}}

\affil[2]{\orgdiv{Department of Psychiatry and Psychotherapy}, \orgname{ University of Regensburg}, \orgaddress{\street{Universitaetsstrasse 84}, \city{Regensburg}, \postcode{93053}, \state{Bavaria}, \country{Germany}}}


\abstract{Clinical records frequently include assessments of the characteristics of patients, which may include the completion of various questionnaires. These questionnaires provide a variety of perspectives on a patient's current state of well-being. Not only is it critical to capture the heterogeneity given by these perspectives, but there is also a growing demand for developing cost-effective technologies for clinical phenotyping. Filling out many questionnaires may be a strain for the patients and therefore costly. In this work, we propose COBALT - a cost-based layer selector model for detecting phenotypes using a community detection approach. Our goal is to minimize the number of features used to build these phenotypes while preserving its quality. We test our model using questionnaire data from chronic tinnitus patients and represent the data in a multi-layer network structure. The model is then evaluated by predicting post-treatment data using baseline features (age, gender, and pre-treatment data) as well as the identified phenotypes as a feature. For some post-treatment variables, predictors using phenotypes from COBALT as features outperformed those using phenotypes detected by traditional clustering methods. Moreover, using phenotype data to predict post-treatment data proved beneficial in comparison with predictors that were solely trained with baseline features.}




\maketitle

\section{Introduction}\label{introduction}

%textcolor{yellow}{Tinnitus can be described as a perception of a sound that cannot be explained by external factors \cite{Eggermont2004}. There have been numerous studies on tinnitus \cite{Eggermont2004, Cima2012, Cima2014}, but due to the heterogeneity of its symptoms there is no standard treatment that fits all patients. Tailored treatments are therefore essential to tackle this issue \cite{Schlee2021}. Phenotyping enables this by capturing heterogeneity and identifying subgroups of patients with comparable traits.}

Clinical records contain a wealth of characteristics of the patients: their vital signs, for example, and other subjective markers determined by questionnaires. There have been numerous studies on tinnitus \cite{Eggermont2004, Cima2012, Cima2014}, but due to the heterogeneity of its symptoms there is no standard treatment that fits all patients. Tailored treatments are therefore essential to tackle this issue \cite{Schlee2021}. Phenotyping enables this by capturing heterogeneity and identifying subgroups of patients with comparable traits. 
For some chronic diseases like tinnitus, it is common that patients are asked to complete a series of questionnaires to assess the severity of their symptoms and its impact on their well-being. However, filling out multiple questionnaires can be a burden for patients \cite{Rolstad2011}, which might affect their adherence to filling out crucial questionnaire items. The cost-effectiveness of digital clinical phenotypes is already seen as a high priority, according to \citet{Huckvale2019}. 

As a response, we propose an algorithm that builds a multi-layered network (MLN) of patient features (assessments, questionnaire scores) and derives communities from it, taking missingness and diversity into account. 
%create a cost-sensitive model for phenotyping (COBALT) and test it using questionnaire data from chronic tinnitus patients. 
We opt for
%We use 
a MLN representation of the patient data, because MLNs can
%This multi-dimensional network 
capture intra- and inter-associations between features, highlighting the heterogeneity of the various perspectives provided by each feature of the data. We propose an extension of our work in \cite{Puga2021} by including a cost-aware aspect in our model.
%We propose a community detection approach that discovers subgroups (communities) in a multi-layer network. These communities can be interpreted as phenotypes if they group patients that portray similar characteristics. 
Our \cobalt{} (\COBALT{}) has thus a cost component that aims to minimize the number of features (questionnaire data) for the construction of the phenotypes, without compromising its quality. Our overarching research question for \COBALT{} is:

\begin{quote}
\textit{How can we exploit features and inter-relationships among them for the discovery of phenotypes?}
\end{quote}
We refine it as follows:
\begin{enumerate}[]
    \item To what extent can tinnitus patients similarity be represented and captured using a multi-layer network? 
    \item How does our approach with a community detection algorithm in a multi-layer network compare to traditional clustering models?
    \item To what extent does including a cost-sensitive model affect the quality of the phenotypes found?
    \item To what extent do missing nodes impact the quality of the phenotypes found?
\end{enumerate}

To this purpose, we investigate how missing data influence the proposed model, as well as how traditional clustering algorithms compare to our approach. We also carry qualitative and quantitative evaluations to ensure that we analyze the impact of our findings with respect to a practical application. More specifically, we build a post-treatment data predictor that learns with baseline features (age, gender and pre-treatment data) and with the phenotypes discovered. We evaluate our approach on a clinical dataset with pre-treatment and post-treatment records of tinnitus patients.

The remainder of the paper is organized as follows: Section \ref{relwork} presents related work on phenotyping and its applications on multi-layer networks, Section \ref{subsec:uhregData} presents the dataset characteristics used in the experiments, Section \ref{methodology} describes the proposed algorithm and Section \ref{evaluation_design} describes how the proposed methodology was evaluated. In Section \ref{results} the results are shown and discussed and Section \ref{conclusion} summarizes the main findings.

\section{Related work}\label{relwork}

The role of cost-awareness in phenotyping research varies among the medical domains. Current tinnitus phenotyping research is centered on clustering methods such as hierarchical clustering analysis and latent clustering analysis \cite{Subphenotypes}. There are many surveys on clinical phenotyping \cite{Subphenotypes, Richesson2016, genes}, but they do not emphasize the issue of feature acquisition cost when constructing phenotypes. 
\citet{Huckvale2019}, on the other hand, identify a set of priorities for the future of clinical digital phenotyping, and they mention the cost-effectiveness of the models as important  while implementing digital phenotyping in a clinical environment.
\citet{Liang2019} also states the importance of cost-effective methods and, more specifically, with respect to the acquisition of data.

\subsection{Phenotyping with MLNs}
Regarding the use of MLNs in the representation of phenotype and genotype data, \citet{Lee2020} review computational methods using MLNs to represent the hierarchy of biological systems. They focus on the quality of the representation of interactions between phenotype data, gene data and SNP data in biological systems, rather than finding phenotypes. More recently, \citet{Yang2021} use functional clustering methods to capture MLNs from any dimension of their genetic data. Concerning community detection for phenotype discovery, \citet{Kramer2020} use the Leiden and Louvain algorithms to find phenotypes using a KNN and CoNet representation of the data. We also use the Leiden algorithm for community discovery in \COBALT{}, but with a different representation of the data. This method discovers a set of communities over all the MLN layers. Hereafter, we use the terms \textit{`set of communities'} and \textit{`partition (over the MLN)'} interchangeably. An overview of community discovery algorithms for MLNs follows after explaining graph pruning.

\subsection{Graph pruning in MLNs}\label{graph_pruning_relwork}
%It is recognized that
%Network simplification improves computation of methods that do not perform well with large networks \cite{Interdonato2020}. 
\citet{Interdonato2020} characterize graph pruning as a network simplification filtering technique. This technique improves computation for methods that do not perform well in large networks. Pruning in networks/graphs is a dimensionality reduction approach that assists in the removal of noisy and redundant edges or nodes. These methods can be grouped into: (i) centrality-based, (ii) node-layer relevance-based and (iii) model-based. The first two focus on node removal, wile the last one focuses on edge pruning. In \COBALT{},
%In our study,
we use the Maximum Likelihood Filter (MLF) of \citet{Dianati2016}, which is mentioned in \cite{Interdonato2020} as an unbiased edge-filtering method that uses maximum entropy.

The main idea behind MLF is to create a null model which generates a ``realized graph" with the same total weight and degree sequence as the original graph and use it to compute a p-value of each edge. Then, edges with p-values above a threshold $\alpha$ (significance level) are pruned. In the realized graph, the nodes are kept the same as in the original graph and with the same degree sequence. Edges are then randomly assigned to a pair of nodes, but with a likelihood. A node with a high degree is more likely to be assigned an edge than a node with a lower degree. This likelihood is computed using the binomial distribution, as follows:

\begin{equation}\label{pruning}
    Pr(\sigma_{ij} = m \rvert k_{i}, k_{j}, E) = \binom{E}{m}p^m(1-p)^{E-m}
\end{equation}

where $m$ is an edge from the set of edges $E$ that connected the pair of nodes $(i,j)$; $k_i, k_j$ are the degrees of nodes $i$ and $j$, respectively; $\sigma_{ij}$ is the weight of the undirected edge between $i$ and $j$, $p=\frac{k_{i}k_{j}}{2E^2}$ and $E = \frac{1}{2}\sum_{i}k_i$.
The p-value of an edge connecting nodes $i$ and $j$ with a weight of $w_{ij}$ is then denoted as $s_{ij}(w_{ij})$. Equation \ref{pruning_pvalue} shows the calculation of the p-value of an edge with weight $w_{ij}$.

\begin{equation}
\label{pruning_pvalue}
    s_{ij}(w_{ij})= \sum_{m \geq w_{ij}}Pr(\sigma_{ij}=m \rvert k_{i}, k_{j}, E)
\end{equation}

\subsection{Community detection
%algorithms for
in MLNs}

%Many complex systems cannot be accurately represented by a single-layer network due to the fact that units may have multi-dimensional properties which cannot be summarized into one node and one type of edge.To capture the intrinsic complexity of these systems, a MLN architecture is required \cite{Huang2020}. However, traditional community detection algorithms are not directly applicable to a MLN.

\citet{Huang2020} highlights the following approaches for the detection of communities: the variational bayes \cite{Ali2019} (aggregation method), GenLouvain \cite{lucas2011} (direct method), Aggregation Pan \cite{Pan2018} (flattening method), the ParticleGao \cite{Gao2019} (flattening method) and the multilayer label propagation \cite{Alimadadi2019} (MNLPA, direct method) algorithms.

\subsubsection{Variational Bayes with stochastic block model}
This method by \citet{Ali2019} proposes to detect shared and unshared communities in a multiplex network. The nodes are represented in a community wise connectivity matrix in blocks. This matrix maps the probability of the nodes being connected to other nodes. Nodes are placed in the same block if their edges are stochastically similar. The Poisson distribution is used to calculate the probability of the nodes being connected.

The 
%determination of the 
optimum number of blocks that represent the nodes is determined
%done
via the Bayes factor \cite{Ali2019} and
%. After that, 
communities are
%can be 
extracted accordingly. \citet{Ali2019} formulate the Weighted Stochastic Block Model (WSBM) adapted to a multiplex network. However, it is not stated if the achieved results outperform a benchmark method like Louvain.

\subsubsection{Modularity optimization with GenLouvain}
The GenLouvain algorithm is a modularity optimization algorithm that is widely used and considered a benchmark algorithm \cite{Huang2020}. 

%explain modularity
Modularity is a metric that evaluates the quality of the partition of the network into communities. The multislice modularity is a modularity metric adapted to MLNs introduced by \citet{Mucha2010}. This metric can be defined as follows:
\begin{equation}\label{modularity_equation}
Q_M = \sum_{ij\alpha\beta}\frac{(A_{ij\alpha} - \gamma\frac{k_{i\alpha}k_{j\alpha}}{2m\alpha})\delta_{\alpha\beta} +  \delta_{ij}C_{j\alpha\beta}}{2\mu}\delta(g_{ij}, g_{j\beta})
\end{equation}

where $\mu$ is the number of edges in the MLN, $\gamma$ is a resolution parameter, $A_{ij\alpha}$ is the value of the edges between $i$ and $j$ in the layer $\alpha$, $C_{j\alpha\beta}$ is an inter-layer edge between the same node $j$ that belongs to layer $\alpha$ and $\beta$ and $k_{i\alpha}$ represents the degree of node $i$ in the layer $\alpha$. 

The main difference between the GenLouvain algorithm and the traditional Louvain algorithm is that the modularity metric is replaced by the multislice modularity \cite{lucas2011}.
Despite of Louvain being one of the benchmark algorithms for community detection based on modularity optimization, it can lead to sub-optimal partitions as shown by \citet{Traag2019}. To tackle this, \citet{Traag2019} present the Leiden algorithm as an improvement to the Louvain algorithm. 
They include a refinement phase in which the nodes to be moved to another community do not have necessarily to provide the highest increase on the quality function. This is a major distinction to Louvain, where the approach is greedy and nodes are only assigned to another community if this results in the largest increase of the quality function. Nonetheless, in the Leiden algorithm there is a likelihood associated with the decision on which nodes should be moved to other communities. The higher the increase in the quality function provoked by moving a node, the higher the likelihood of that node to be selected.

\subsubsection{AggregationPan} \citet{Pan2018} propose an algorithm that aggregates the edge weight matrices and then applies a cut-off, so that the edge weights with low values ($<\tau$) are converted to 0. However,
%This is an 
aggregation 
%method, which 
is criticized in the literature since its simplification may mask the true nature of the initial modular patterns \cite{Huang2020}. Plus, there is the need to define $\tau$.

\subsubsection{Particle competition} \citet{Gao2019} work is based on a particle competition algorithm for MLNs. The fundamental concept is to insert a certain amount of K particles into network nodes. Each particle's goal is to dominate as many nodes as possible while also safeguarding their current dominated nodes from other particles. When a particle visits a node, it gains strength while weakening the other particles in the node. At the end of the algorithm, each particle should represent a community. The particles can move in two ways: random walking and preferential walking. Random walking chooses a node at random from its neighbors, whereas preferential walking visits a neighbor with a high dominance. A balancing parameter is then used to balance these two types of walking. This means that a node's propensity to choose a random or preferential walk is affected by this value. The experiments were, however, not tested in real world networks.

\subsubsection{MNLPA} This algorithm is introduced by \citet{Alimadadi2019} and consists in a generalization of the label propagation algorithm (LPA) to multiplex networks. At the beginning, each node is assigned a label. Then, similarity measures are used to compare nodes. If these two nodes have a certain similarity metric higher than a given threshold, then the label of the two nodes is replaced by a common label to both. This is done until the stopping criteria are achieved.
However, the survey from \cite{Huang2020} states that this method is better suited for directed and weighted networks rather than for general MLNs. It is also mentioned that it might be unstable due to the threshold parameter defined, which impacts strongly the network partition.

\subsection{Similarity functions for MLNs}
\citet{Brodka2018} presents some methods to compare distributions between layers in a multiple network: (i) dissimilarity index, (ii) Kullback-Leibler, (iii) Jensen-Shannon and (iv) Jeffrey.

For comparing properties with binary and numeric values, some other measures are mentioned by \citet{Brodka2018}. For properties with binary values: Russel-Rao, Jaccard, coverage, Kulczyński, simple matching coefficient (SMC) and Hemann. For properties with numeric values, the following metrics are normally used: cosine similarity, Pearson correlation similarity, Spearman correlation coefficient.

\subsubsection{Comparing the networks}
\citet{Tantardini2019} present approaches for comparing networks, both when the nodes are aligned and the pairs are known (known-node correspondence, KNC) and when the networks comprise nodes that are not aligned and hence distinct (unknown-node correspondence, UNC).
For undirected and weighted networks, some methods are presented in Table \ref{tab:net_similarity_techniques}.

\begin{table}[htb]
\resizebox{\columnwidth}{!}{%

\centering
\begin{tabular}{ll}
\cline{1-2}
\textbf{Network type} & \textbf{Techniques}                         \\ \cline{1-2}
\hspace{7pt}KNC          & Euclidean, Manhattan, Canberra distances,  \\
                         & Weighted Jaccard distance (WJAC)            \\\cline{1-2}
\hspace{7pt}UNC                      & Global statistics                           \\
                         & Spectral adjacency, \\
                         & Laplacian SNL distances,          \\
                        &  MI-GRAAL                                      \\
                         & NetLSD                                      \\
                         & Portrait divergence                         \\ \cline{1-2}
\end{tabular}%
}
\caption{Network similarity techniques for KNC and UNC}
\label{tab:net_similarity_techniques}
\end{table}
Concerning UNC, it is stated that global statistics are not a reliable tool for comparing layer similarity because they are overly simplistic.
Spectral approaches also have several downsides, such as co-spectrality between graphs, dependence on matrix representation, and abnormal sensitivity.
According to \citet{Tantardini2019}, the MI-GRAAL has a significant computational cost, but the Portrait divergence is largely efficient with small to medium graphs.

\subsubsection{Comparing communities}
All of the above metrics focus on describing the similarity between networks based on their edges, nodes and/or other properties. For the current work, we are interested in comparing a specific property - the community structures among layers. Recently, \citet{Ghawi2021} identified some drawbacks when using extrinsic (that require ground truth) evaluation metrics, such as the direction of the comparison. More precisely, if we have 2 layers of a MLN, $l_{\alpha}$ and $l_{\beta}$, there are two-ways for matching the communities in each one of them. One can use either $l_{\alpha}$ or $l_{\beta}$ as the basis of the comparison and this may lead to different results. For instance, let $C^{l_{\alpha}} = \{c_1, c_2, ..., c_r\}$ be the set of communities in $l_{\alpha}$ and $C^{l_{\beta}} = \{c_1, c_2, ..., c_k\}$ be the set of communities in $l_{\beta}$. Let $n_{C_1}^{l_{\alpha}} = \{n_1, n_2, ..., n_m\}$ be the set of nodes in $c_1$ of layer $l_{\alpha}$ and $n_{C_1}^{l_{\beta}} = \{n_1, n_2, ..., n_n\}$ the set of nodes in $c_1$ of layer $l_{\beta}$. In these communities, if we consider $l_{\alpha}$ as the basis of the comparison (the ground truth) and $l_{\beta}$ as the layer to be compared with the group truth. If we look at node existence, we will check for the number of nodes from $n_{C_1}^{l_{\alpha}}$ that are in $n_{C_1}^{l_{\beta}}$. In the case that $n_{C_1}^{l_{\alpha}}$ has 15 nodes and $n_{C_1}^{l_{\beta}}$ has 30, from which 15 are the same as in $l_{\alpha}$, then the similarity metric gives a perfect similarity. Therefore, a two-way matching is required, in which each layer is considered the ground truth - one at a time - and then both similarity values are combined into a single metric. Since purity and F-measure are among the most commonly used metrics for clustering evaluation and they allow the comparison between two clustering solutions, they fit the purpose of the current work.

% explain purity
\citet{Ghawi2021} convert the purity in a two-way matching by computing firstly the purity of $l_\alpha$ against $l_\beta$, $purity^{[l_\alpha \| l_\beta]}$, and vice-versa, $purity^{[l_\beta \| l_\alpha]}$. The harmonic mean between both values is computed and the final purity is as follows:

\begin{equation}
purity = \frac{2*purity^{[l_\alpha \| l_\beta]}*purity^{[l_\beta \| l_\alpha]}}{purity^{[l_\alpha \| l_\beta]}+purity^{[l_\beta \| l_\alpha]}}
\end{equation}

With this metric, we can study the extent to which the clustering is ``pure," with respect to node existence.

% explain f-measure
F-measure is computed using the recall and precision of the clusters. The precision of a cluster is the same as its purity. The recall metric evaluates the fraction of nodes that are shared between the ground truth cluster and the system-generated one.
In the end, the F-measure of a cluster is the harmonic mean of its precision and recall. Considering that $F^{[l_\alpha \| l_\beta]}$ is the F-measure when considering $l_\alpha$ as ground truth and $l_\beta$ as the system-generated clustering solution and $F^{[l_\beta \| l_\alpha]}$ is the F-measure of the clustering solution when $l_\beta$ is considered as ground-truth and $l_\alpha$ as the system-generated clustering solution. The overall F-measure is given by the harmonic mean of both values:

\begin{equation}
F = \frac{2*F^{[l_\alpha \| l_\beta]}*F^{[l_\beta \| l_\alpha]}}{F^{[l_\alpha \| l_\beta]}+F^{[l_\beta \| l_\alpha]}}   
\end{equation}

\section{Materials}
\label{subsec:uhregData}
The data analyzed in this study refers to chronic tinnitus patients admitted to the University Hospital of Regensburg. The data was gathered between January 3, 2016 and May 28, 2020.
The studies involving human participants were reviewed and approved by the ethics committee of the University Regensburg. The patients/participants provided their written informed consent to participate in this study.

At the time of admission, each patient fills out a series of questionnaires meant to assess some of the patient's mental and physiologic symptoms. The questionnaire data used in this research was gathered from five questionnaires: tinnitus questionnaire by Goebel and Hiller (TQ) \cite{tq}, tinnitus handicap inventory (THI) \cite{thi}, tinnitus functional index (TFI) \cite{tfi}, major depression inventory (MDI) \cite{mdi}, and tinnitus impairment questionnaire (TBF12) \cite{tbf12}. In total, data from 1087 patients were considered.

Two time points are considered: $t_0$ denotes the so-called `screening', where all questionnaires are answered and the treatment is scheduled to start; $t_1$ denotes the moment of the last visit of the patient at the end of the treatment, whereby some of the questionnaires are answered again and the scores are compared. We also use the expressions `pre-treatment' and `before treatment' (moment) for $t_0$ and `post-treatment', `after treatment' (moment) for $t_1$.

Table \ref{tab:materials_table} shows the number of patients with available data at $t_0$ and the number of patients with records at both $t_0$ and $t_1$. The last column shows the ranges of the questionnaire scores. It can be seen that the questionnaires have different value ranges. However, for all of them smaller values are better, in the sense that the patient is in better health. 

\begin{table}[!ht]
\centering
\begin{tabular}{lcll}
\cline{2-4}
                                      & \multicolumn{1}{l}{$t_0$} & $t_0$ and $t_1$ & Range        \\ \cline{2-4} 
\textrm{THI questionnaire}   & $1067$                      & $123$             & {$[0, 100]$} \\
\textrm{MDI questionnaire}   & $981$                     & $109$             & {$[0, 50]$}  \\
\textrm{TFI questionnaire}   & $798$                       & $87$              & {$[0, 100]$} \\
\textrm{TQ questionnaire}    & $746$                       & $70$              & {$[0, 84]$}  \\
\textrm{TBF12 questionnaire} & $700$                       & $35$              & {$[0, 24]$}  \\ 
\cline{2-4} 
\end{tabular}
\caption{Number of patient records per questionnaire at $t_0$, $t_0$ and $t_1$ and range of each questionnaire -- ordered on the number of questionnaires at $t_0$ descending}
\label{tab:materials_table}
\end{table}

\section{Methodology}\label{methodology}
We adopt an iterative and cost-sensitive approach to find communities in a MLN and name it \COBALT (\cobalt{}).

The first phase is denoted as representation. The nodes, edges and layers are defined in this phase. Then, pruning is applied to remove edges that are not statistically significant.
Subsequently, we introduce the cost-sensitive component to our model. At this phase, the minimum set of layers that capture patient phenotypes without compromising its quality is found. This is done iteratively, with layers being added to the structure in a specific order determined by their cost. The cost of each layer is also modeled. 

Subsection \ref{representation} describes the representation phase, subsection \ref{methodology_pruning} describes the graph pruning phase, subsection \ref{layer_cost_model} presents the layer cost model, subsection \ref{methodology_illustration} illustrates an example of our MLN, subsection \ref{search_algorithm} describes the search algorithm and subsection \ref{stopping_criterion_methodology} describes the stopping criteria for the proposed algorithm.

\subsection{Representation}\label{representation}
Let $p \in \mathcal{P}$ be the set of nodes (for our application: patients) in a graph $g \in G$ (for our application: questionnaires). Each $g$ is considered a layer and therefore we denote it instead by $l \in L$. A node $p_i$ in layer $l$ is denoted as $(p_i, l)$. 
Let $e \in E$ denote the edges in layer $l$. The edge between two nodes $p_i$ and $p_j$ in layer $l$ can be defined as $e_{(p_i, p_j, l)}$, which we simplify to $(p_i, p_j, l)$.
We denote the edge weight (which is an attribute of the edge) between these nodes as $w_{p_{i}, p_j, l}$.

A MLN is constructed to combine the multiple data features into a single structure. This network has $L$ layers and each layer has nodes and connections between them (edges) with a weight associated to it (edge weight). 

We span an ``intra-layer" edge between each pair of nodes within the same layer, assigning the value of the normalized distance between the nodes in this layer to it (edge weight). We span an ``inter-layer" edge across two layers, connecting nodes that represent the same patient in different layers, and assign a value to it based on the distance between the nodes.

When building the inter-layer edges, two types of edges may exist:
\begin{enumerate}
    \item edges of a node with itself between two layers
    \item edges between different nodes located in different layers
\end{enumerate}
We generate the inter-layer edges to incorporate the different perspectives given by different features of a node. Hence, only inter-layer edges between the same nodes in different layers are incorporated.
We use the same logic as in \cite{Puga2021}, which we explain in more detail in \ref{subsub_intra_layer} and in \ref{subsub_inter}.

\subsubsection{Intra-layer edges}\label{subsub_intra_layer} 
The distances between nodes within a layer are represented by intra-layer edges. A layer represents a feature in our algorithm and in the specific case of our application it represents a questionnaire. These, however, can be generalized to other types of numerical features.

Considering two nodes $p_{i}$ and $p_{j}$ in the same layer, the edge between them are defined by the difference between their feature values, which in our case are questionnaire scores.
Assuming a questionnaire $l \in L$ ($L$ is the set of layers that represent questionnaires), $score_{p_{i}, l}$ and $score_{p_{j}, l}$ denote the scores of nodes $p_i$ and $p_j$ in layer $l$, respectively.

The larger the discrepancy in scores between nodes, the lower their edge weight. This ensures that their connection is represented by a ``weak" edge weight if their scores are not close. To accommodate for this, we define the weight as in Equation \ref{intra_layer_weight}, which describes the transformation $1/x$.

\begin{equation}\label{intra_layer_weight}
    w_{p_{i},p_{j},l} = \frac{1}{\abs{score_{p_{i}, l}' - score_{p_{j}, l}'}}
\end{equation}

In Equation \ref{intra_layer_weight}, $w_{p_{i},p_{j},l}$ denotes the edge weight that connects $p_{i}$ and $p_{j}$ in layer $l$.
The scores are then normalized by subtracting the mean and dividing by the standard deviation.

% The scores of each node vary according to a known range of possible values. The mean and standard deviation are thus used to calculate the standardization of the difference in scores between nodes. The mean and standard deviation of the questionnaire scores of a questionnaire represented by layer $l$ are denoted by $\mu_l$ and $\sigma_l$, respectively. This phase guarantees that the inter-layer edges take into account the existence of different value distributions among questionnaires. \textcolor{red}{ As a result, scores are standardized as indicated in Equation \ref{transformation_mean_std}.}

% \begin{equation}\label{transformation_mean_std}
%     score_{p_{i}, l}' = \frac{score_{p_{i}, l} - \mu_l}{\sigma_l}
% \end{equation}

\subsubsection{Inter-layer edges}\label{subsub_inter}

Equation \ref{coupling_edges_weight} shows how to compute the weight $w_{(p_{i}, l_{\alpha}), (p_{i}, l_{\beta})}$ of an inter-layer edge that connects $(p_i, l_{\alpha})$ and $(p_i, l_{\beta})$.

\begin{equation}
\label{coupling_edges_weight}
     w_{(p_{i}, l_{\alpha}), (p_{i}, l_{\beta})} = \frac{1}{\abs{score_{p_{i}, l_{\alpha}}' - score_{p_{i}, l_{\beta}}'}}
\end{equation}

$score_{p_{i}, l_{\alpha}}'$ and $score_{p_{i}, l_{\beta}}'$ denote the normalized score of layer $l_{\alpha}$ and $l_{\beta}$ for node $p_{i}$, respectively. $l_{\alpha}, l_{\beta} \in L$ correspond to layers $\alpha$ and $\beta$, respectively.

\subsection{Graph pruning}\label{methodology_pruning}
The MLN representation of the previous paragraphs results into a fully-connected network at each layer. To ensure that only the most important edges are retained, we apply next a 
%This leads not only to high computational efforts, but it may also affect the quality of community detection algorithms and the discovery and analysis of inherent structures to the network. We include 
graph pruning step for inter-layer and intra-layer edges, using
%phase in our workflow and use 
the maximum likelihood filter (MLF) proposed by \citet{Dianati}, see subsection \ref{graph_pruning_relwork}.
%as a method to prune both inter- and intra-layer edges.
We fix the threshold for pruning the edges at $0.05$, i.e. edges with a p-value higher than $0.05$ are removed.

\subsection{Cost model for layer selection} 
\label{layer_cost_model}

Certain questionnaires may be more likely to be completed than others (due to the type of questions). This aspect should be considered while deciding on the next layer to be added to the MLN. 
The similarity of communities between layers is a second criterion of relevance. We intend to add a layer containing additional information about the patients while avoiding adding redundant information. As a result, the more distinct the layers are with respect to their communities, the less redundant they are. The goal is to add a layer with a low community similarity.

Hence, the cost of a layer is calculated using two terms: an availability ratio term and a community similarity term. We formulate the function to measure the cost of a layer $l_\alpha$, $C_{l_{inc}, l_{\alpha}}$, with respect to the incumbent set of layers in the network ($l_{inc}$) as:
\begin{equation}
\label{eq:layer_importance}
    C_{l_{inc},l_{\alpha}} = \frac{1}{A_{l_{inc}, l_\alpha}} + CS_{l_{inc}, l_{\alpha}}
\end{equation}
where $A_{l_{inc}, l_\alpha}$ denotes the availability ratio term and $CS_{l_{inc}, l_{\alpha}}$ the community similarity term. These two terms are described hereafter.

\subsubsection{Availability ratio term}
We term the ratio of completion of the questionnaires as availability ratio. $A_{l_{inc}, l_\alpha}$ denotes the availability ratio of the questionnaire that is represented by layer $l_\alpha$ against $l_{inc}$. It is the ratio of nodes that are not missing in $l_\alpha$ from the nodes that are already in $l_{inc}$.

Considering the set of nodes in layer $l_\alpha$ as $p^{l_\alpha} = \{p_1, p_2, ..., p_m\}$ and in the incumbent layer or layer set, $l_{inc}$, as $p^{l_{inc}} = \{p_1, p_2, ..., p_n\}$, $A_{l_{inc} l_\alpha}$ is given by:
\begin{equation}
\label{eq:availabilityRatio}
    A_{l_{inc}, l_\alpha} = \frac{\abs{p^{l_\alpha} \cap p^{l_{inc}}}}{\abs{p^{l_{inc}}}}
\end{equation}

The higher the amount of missing nodes in a layer, the fewer the number of nodes (patients) that are added to the MLN. The goal is to add the maximum information about the nodes in each iteration. Therefore, we aim to maximize the availability ratio term. 

\subsubsection{Community similarity term}
We define the community similarity term as a measure of how similar are communities between layers, with respect to the assigned nodes. More specifically, we focus on quantifying the shared nodes between communities of different layers. 

We use the bi-directional F-measure to quantify this term.
There are two reasons for this choice: (i) the metric includes both the purity/precision and the recall of the solution and (ii) it handles the absence of ground truth by considering as ground truth one layer at a time.

The community similarity between layers $l_\alpha$ and the incumbent layer or set of layers $l_{inc}$ is modeled as:

\begin{equation}
\label{eq:communitySimilarity}
    CS_{l_{inc},l_{\alpha}} = \frac{2*F^{[l_{inc} \| l_\alpha]}*F^{[l_\alpha \| l_{inc}]}}{F^{[l_{inc} \| l_\alpha]}+F^{[l_\alpha \| l_{inc}]}} 
\end{equation}

As previously stated, we intend to use layers that provide the maximum additional information about the nodes. If the community structure of two layers is substantially similar, they are seen as redundant. We intend the exact opposite: to add layers with a community structure that differs from that of the incumbent (current) network. As a result, $CS_{l_{inc},l_{\alpha}}$ should be minimized.

\subsection{Illustration example}\label{methodology_illustration}
\begin{figure}[!ht]
\centering
       \includegraphics[height=6.5cm,keepaspectratio]{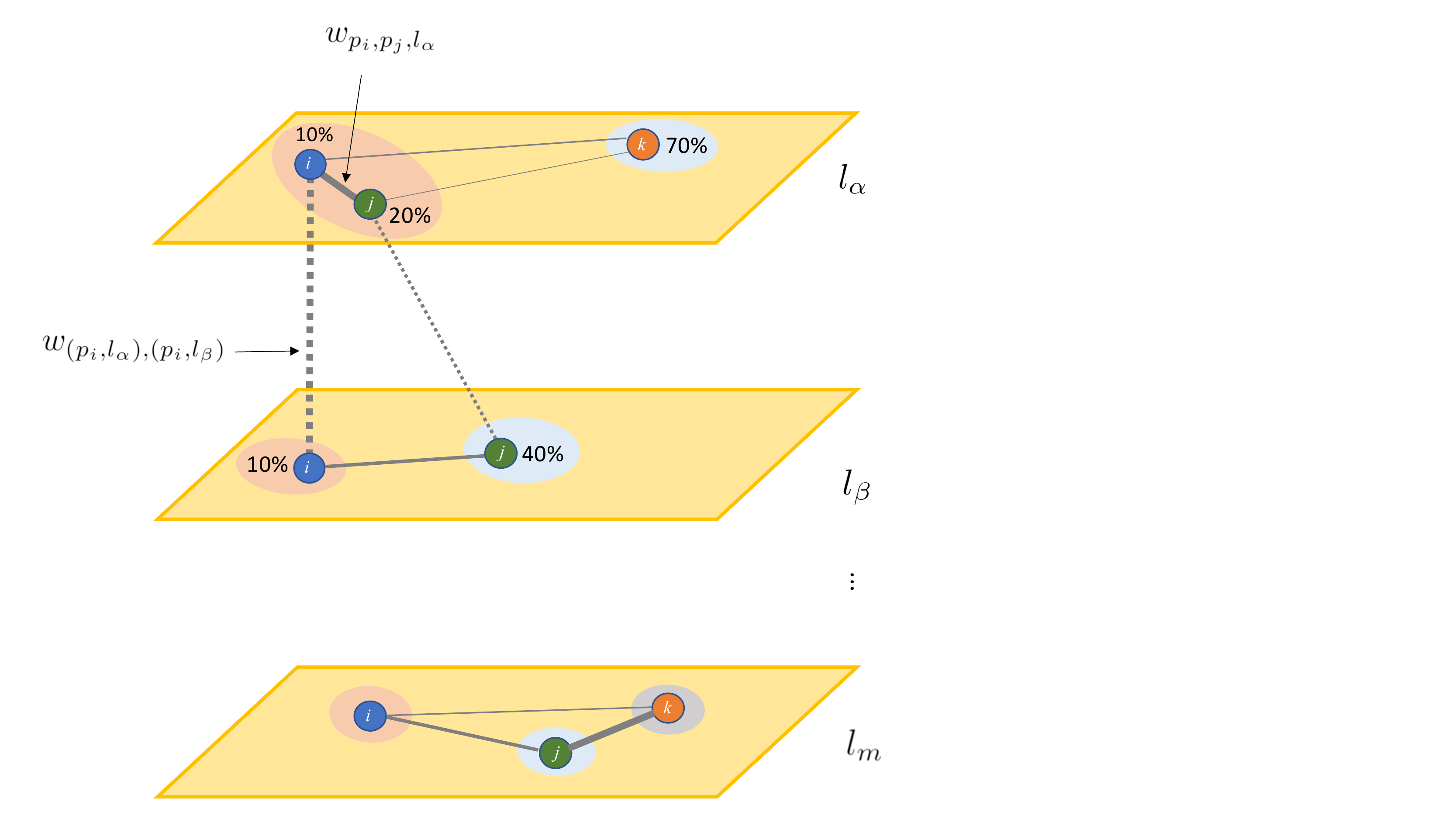}
\caption{Structure of a MLN with communities generated in each layer.}
\label{fig:representation_network}
\end{figure}

Figure \ref{fig:representation_network} illustrates the structure of a MLN with communities. Thicker edges between nodes correspond to a high similarity between them. 
Nodes are represented by dots of different colors, and they may be present in all or only a subset of the layers. Communities are represented by colored circles that surround one or more nodes.

In Figure \ref{fig:representation_network}, the edge that connects nodes $p_i$ and $p_j$ in layer $l_{\alpha}$ is thicker than the ones that connect them to $p_k$ in the same layer. This is because these two nodes are more similar between themselves than with $p_k$. The score of $p_i$ is $10\%$ whereas the score of the $p_j$ is $20\%$ in $l_{\alpha}$. Hence, the difference between them, in layer $l_{\alpha}$, is only $10\%$. If we compare the score of $p_k$ (of $70\%$), this difference is higher and hence the similarity between $p_i$ and $p_j$ with $p_k$ is lower. The edges are therefore represented by thinner lines.

Since the layers represent different features, the similarity between the same set of entities may differ from layer to layer. An example is illustrated in layers $l_{\alpha}$ and $l_{\beta}$, where the pair of nodes $p_i$ and $p_j$ have a higher distance between them in $l_{\beta}$ than in $l_{\alpha}$, making the connection between them ``weaker".
One of the motivations to use MLNs is the fact that inter-layer edges provide information about the association between features (layers). This association is taken into account when building the communities with a community detection algorithm, by using both intra-layer distances and inter-layer distances for modeling.
%, but also the inter-layer distances.

The example also illustrates communities generated in each layer, individually. Figure \ref{fig:representation_network} represents a scenario in which a community detection algorithm runs on each single-layer network. In the example, node $p_i$ appears in all layers and in the same community, but node $p_j$ is assigned to the red community in $l_{\alpha}$ and to the blue community in $l_{\beta}$ and in $l_{m}$. Additionally, node $p_j$ is assigned to the same community as $p_i$ in $l_{\alpha}$, but to other communities in the other layers. We assume that if the community structure of layers are heterogeneous, then they provide different information about the nodes (patients). These three layers could then be considered as not redundant with respect to its community structure.

As previously mentioned, using a bi-directional metric for layer similarity is crucial in the context of the current work, due to the fact that ground-truth is unknown. 
For instance, if we compare the red community in $l_{\alpha}$ to the red community in $l_{m}$ and use the $l_{\alpha}$ as the ground truth, then its precision and recall is $1$ and $\frac{1}{2}$, respectively. For the blue community, precision and recall are both $0$. If we compute the weighted sum of the precision and recall, then this results in $\frac{1}{2}$ of precision and $\frac{1}{4}$ of recall. Therefore, the F-measure ($F^{[l_{\alpha} \| l_m]}$) over all communities is $\frac{1}{3}$.
In contrast, when we consider $l_m$ as the ground-truth, then the precision and recall for the red community are, respectively, $\frac{1}{2}$ and $1$. For the blue community, both values are $0$. In this case, we also include the grey community and compute its precision and recall, which is $0$ for both values. The weighted average of the recall, for this community solution, is $\frac{1}{6}$ and of the precision is $\frac{1}{3}$. The F-measure ($F^{[l_{m} \| l_{\alpha}]}$) is then $\frac{2}{9}$. As a result, there is a significant difference when one layer is assumed to be the ground-truth solution versus the other.

\subsection{\cobalt{} -- \COBALT{}} \label{search_algorithm}
%We identify two main objectives that support the need for the search algorithm:
%to build a MLN that maximizes community modularity (see Equation \ref{modularity_equation}) while minimizing the cost of the layers being added (Equation \ref{eq:layer_importance}). 
On the basis of the aforementioned cost model, we now present 
our algorithm \COBALT{} for cost-based layer selection and community construction.

Table \ref{tab:notation table} displays a description of the main variables used in the pseudocodes that follow.
\begin{table}[htb]
\centering
\begin{tabular}{lp{0.6\columnwidth}}
\hline
\textbf{Symbol} & \textbf{Description} \\ \hline
$G$               & set of graphs  \\
%\hline
$\mathcal{S}_{single}$  & set of the community memberships (partitions) of the nodes for each of the single-layer networks $g\in{}G$
\\
$g_{bestsingle}$           & first graph chosen for the MLN, it contains the partition with the best modularity \\
$S_{bestsingle}$ & partition of the first graph \\
$\mathcal{L}$       &  the network structure: a single-layer or multi-layer network
\\
\hline
\end{tabular}
\caption{Notation table}
\label{tab:notation table}
\end{table}

Algorithm \ref{start_solution} shows the pseudocode for the generation of the starting solution.
\begin{algorithm}
\caption{
\textbf{\COBALT{} Initialization:}
%\textbf{Initial Solution}: 
Pseudo code for choosing the first layer}
\label{start_solution}
\begin{algorithmic}[1]
 \renewcommand{\algorithmicrequire}{\textbf{Input:}}
 \renewcommand{\algorithmicensure}{\textbf{Output:}}
 \Require $G$
 \Ensure $g_{bestsingle}$  
   \State $\mathcal{S}_{single} \gets \emptyset$
   \State $g_{bestsingle} \gets \mathit{NULL}$ 
   \State $S_{bestsingle} \gets \mathit{NULL}$; $q_{bestsingle} \gets -\infty$
%   \State $\mathcal{S}, Q, N \gets \{ \emptyset, \emptyset, \emptyset \}$
  \For {$g \in G$}
    %\State $n_{incumbent} \gets g$
  \State $(S,q) \gets  \mathit{Leiden}(g)$
  \State $\mathcal{S}_{single} \gets \mathcal{S}_{single} \cup \{S\}$
  \If {$q_{bestsingle} < q$}
  \State $g_{bestsingle} \gets g$ 
  \State $S_{bestsingle} \gets S$ ; $q_{bestsingle} \gets q$
  %\State 
  \EndIf
  %\par\hskip\algorithmicindent and $n_{inc}$ to $N_{all}$\par
  \EndFor \\
%  \State $g_{bestsingle} \gets$ $g$ with $\max(Q)$\\
  \Return $g_{bestsingle}, S_{bestsingle}, \mathcal{S}_{single}$

% \Return $g_{bestsingle}, S_{single}, N_{all}$
\end{algorithmic}
\end{algorithm}
% explain the algorithm
The input of Algorithm \ref{start_solution} is the set of all single-layer graphs $G$. 
%that is constituted by one graph per layer. 
For each $g\in{}G$, \COBALT{} invokes the Leiden algorithm 
%is executed and 
which builds a set of communities and returns two objects:
%are discovered. The outputs are of two types: 
the `partition' $S$, which encompasses the community membership of each node for the incumbent graph $g \in G$, and $q$ -- the modularity of 
%the incumbent partition 
$S$. The partition $S_{bestsingle}$ with the largest modularity and the corresponding graph $g_{bestsingle}$ are returned as initial solution of \COBALT{}, together with the set of all single-layer partitions.
%and it forms the output of the algorithm.
%\input{algo1_withQ}

After the initialization, \COBALT{} gradually adds layers to build up a multi-layer network and derive the best set of communities in it. In each iteration, it invokes the cost model
%In the second phase of our algorithm, we extend our work in \cite{Puga2021} and modify the layer selection criterion by replacing it with the cost-based model previously described 
(cf. subsection \ref{layer_cost_model}) to add the least-cost layer, until a stopping criterion is met. The pseudocode is depicted in
%We reformulate the search algorithm as in
Algorithm \ref{global_search}. Note that we consider the first iteration of Algorithm \ref{global_search} as iteration 2, since the iteration 1 represents Algorithm \ref{start_solution}.
%The number of iterations is equal to the number of layers. 

\begin{algorithm}
\caption{
Iterative Layer Selection: Pseudo-code of the subsequent iterations of \COBALT{}
%\textbf{Layer Selection}: Pseudo code of the search algorithm
}
\label{global_search}
\begin{algorithmic}[1]
 \renewcommand{\algorithmicrequire}{\textbf{Input:}}
 \renewcommand{\algorithmicensure}{\textbf{Output:}}
%  \Require $G, g_{bestsingle}, S_{single}, N_{all}$
 \Require $G, g_{bestsingle}, S_{bestsingle},\mathcal{S}_{single}$

 \Ensure $S_{best}$ (best set of communities)
  \State $G_{candidates} \gets G \setminus \{g_{bestsingle}\}$
  \State $S \gets S_{bestsingle}$
  \State $\mathcal{L} \gets g_{bestsingle}$ 
%   \State $n_{inc} \gets n \in N_{all}$ that corresponds to $g_{bestsingle}$
  \While{$G_{candidates} \neq \emptyset$ and \textit{SC} is False}
 
 %    \State $g \gets \argmin_{u\in G_{candidates}}\{cost(S_u,S)\}$ \par
    \State $g\gets{}\mathit{NULL}$; $c\gets{}+\infty$
    \For{$u\in G_{candidates}$}
        \State $c_u \gets cost(\mathcal{L},u,S,S_u)$
        \If{$c_u < c$}
            \State $g\gets{}u$ ; $c\gets{}c_u$
        \EndIf
    \EndFor
    % \For{$g \in G_{candidates}$}
    %     \State $s_g \gets s \in S_{single}$ that corresponds to $g$
    %     \State $cost_{inc, g}\gets$\textit{cost} $(s_{inc}, s_g)$
    % \EndFor
    % \State $g_i\gets g \in G_{candidates}$ with $min(cost)$
    \State extend $\mathcal{L}$ with $g$
    %\State $n_{test} \gets$ add a layer $g_i$ to $n_{inc}$
    \State $(S_{\mathcal{L}}, q_{\mathcal{L}}) \gets Leiden(\mathcal{L})$ 
    %\State $s_{test}, q_{test}, n_{test} \gets Leiden(n_{test})$ 
%   \If{$q_{test} > q_{inc}$}
%     \State $s_{inc} \gets s_{test}$
%     \State $q_{inc} \gets q_{test}$
%     \State $n_{inc} \gets n_{candidate}$
%   \EndIf
   \State  $G_{candidates} \gets G_{candidates} \setminus \{g\}$
  \EndWhile \\
%  \State $s_{final} \gets s_{inc}$\\
 \Return $S_{\mathcal{L}}$
\end{algorithmic} 
\end{algorithm}

The input of Algorithm \ref{global_search} is the set of single-layer graphs $G$ and the outputs of Algorithm \ref{start_solution}. The graph $g_{bestsingle}$ becomes the first layer of the MLN structure $\mathcal{L}$, which is built iteratively by adding layers.
%$\mathcal{L}$ denotes the single- or multi-layer structure of the incumbent iteration. More specifically, this variable is equivalent to $g_{bestsingle}$ in the initial phase of algorithm, but as layers are added, $\mathcal{L}$ is updated to a multi-layer network structure.
%
To expand $\mathcal{L}$, the list of candidate graphs $G_{candidates}$ is created from $G$ and is dynamically updated to remove the graph $g$ added at each iteration. To choose this graph $g$,
%include only graphs that were not added yet to $\mathcal{L}$. No repeated layers can be added and therefore $G_{candidates}$ should be empty at the last iteration of the algorithm. 
%
%Therefore, at the beginning of the algorithm $g_{bestsingle}$ is removed from the set of graphs $G$ and this results into the set of candidates $G_{candidates}$. Then, 
the layer cost is computed for each candidate $u \in G_{candidates}$, using the cost formula of Eq. \ref{eq:layer_importance}.

This computation is made between the candidate $u$ and $\mathcal{L}$, but it demands also the community memberships (for the community similarity term). To make this more clear, we include the corresponding partitions. 
%The cost is computed based on the community similarity between both layers and the availability ratio term.
%\textcolor{red}{Hence, the community memberships are required as inputs: $S_u$ for the candidate and $S_{inc}$ for the incumbent partition. } 

Next, the least-cost candidate $g$ is selected. $\mathcal{L}$ is expanded to incorporate $g$ and the inter-layer edges between its nodes and the nodes in $g$. 
%This consists in the first MLN (a 2-layer network). Modularity of this incumbent solution, $q_{inc}$, is computed. The subsequent iterations follow to the same premise. 
Then, the set of communities is built, $G_{candidates}$ is updated and the next iteration starts.

The algorithm runs while there are candidates in $G_{candidates}$ and while the stopping condition \emph{SC} is false. The stopping condition monitors the cost of the growing MLN, and is specified in the next subsection \ref{stopping_criterion_methodology}.

\subsection{Stopping condition for \COBALT{}} \label{stopping_criterion_methodology}
The main loop of \COBALT{} (cf. Algorithm \ref{global_search}, \textbf{while}-loop) gradually adds each layer; the cost function only decides which layer to add next.
%
%As stopping criterion we use the change of the availability ratio term from one iteration to the next.
Since layers are chosen on the basis of availability ratio \emph{and} community similarity (Eq. \ref{fig:mod_layer_cost}), and since both factors take positive values, cost cannot be negative. Availability ratio may increase or drop from one iteration to the next, though. Two stopping criteria can be derived from it:
\begin{itemize}
    \item \textit{SC1:} \COBALT{} stops when the availability ratio decreases towards the previous iteration
    \item \textit{SC2:} \COBALT{} stops when the availability ratio drops \emph{and} the community similarity increases
    %in comparison to the previous iteration
\end{itemize}
where the `previous iteration' is the 2nd (i.e. the 1st after the initialization) or a later one.

We choose SC1 as stopping condition in the \textbf{while}-loop. In our experiments, SC1 is not used, because we study the behavior of \COBALT{} as each layer is added. We rather report at which iteration \COBALT{} would have stopped, and what would have been the effect on the communities' contribution to predictive power.

% \subsection{Incorporating \COBALT{} into a Phenotype-Aware Treatment Outcome Predictor}
% As pointed out in the Introduction, the main potential of phenotyping is seen in designing more personalized treatments. To assess the merit of our cost-based phenotyping approach towards treatment personalization, we incorporate the communities output by \COBALT{} into an `outcome predictor', i.e. a model that predicts the post-treatment assessment score(s) of a patient given the pre-treatment score(s).

% LOREM IPSUM

%\input{subsection_evaluation}

%\section{Experimental evaluation and discussion}
\section{Evaluation design}\label{evaluation_design}
For the evaluation of \COBALT{}, we use modularity as internal measure and contribution to predictive performance as external measure. We further quantify the impact of missingness.
%We evaluate \COBALT{} on phenotype quality and on sensitivity to missingness. To this purpose, we use data of the outpatient tinnitus clinic described in section \ref{subsec:uhregData}.  
%We evaluate phenotype quality with an internal measure (modularity) and externally as contribution to predictive performance.

%In this section, the results from the experiments are shown. 

\subsection{Phenotype quality as modularity}
\label{subsection_evaluation}
To investigate the role of layer cost on phenotype quality, we use modularity as community quality evaluation measure, cf. formula in Eq. \ref{modularity_equation} and we study how modularity changes as layers are added in a cost-sensitive way.

\subsection{Community visualization scheme}
\label{subsec:FruchtermanReingold}
To acquire insights on how communities change after selecting each additional layer, we use the Fruchterman-Reingold layout \cite{Fruchterman1991} as basis for visualization: nodes that are positioned closer to one another have stronger connections between them than with the ones located far apart. In this layout, we use colors for the communities, i.e. each node takes the color of the community it belongs to. Thus, `good' communities are visualized as graph partitions colored with only one color, while the occurrence of multiple colors in one area of the visualized network indicates that the communities are mixed up.

Since the Fruchterman-Reingold layout is two-dimensional, we place the visualizations of the layers used in each iteration for community detection next to each other; thus, we can see how the colors/communities span across layers.

%Within the scope of our application, however, we want to examine how well the algorithm performs in the domain in which it is tested.

\subsection{Measuring the contribution of phenotypes to predictive quality}
To assess the contribution of the phenotypes discovered by \COBALT{} towards phenotype-sensitive treatment, we first build a `Baseline' set of regressors, each of which predicts
%We do this by training a regressor to predict 
the score of each
%data (at $t_1$), per 
questionnaire (layer) at $t_1$. They use the following features for each patient: age, gender and score of that questionnaire at $t_0$. %before the treatment
%at admission (at $t_0$) 
We compare this `Baseline' to regressors that also exploit community information, namely the ID of the patient's community for each patient/layer node. %to which each patient (node) is assigned to.
Since the patient's community changes as layers are added by \COBALT{}, we train one regressor on the community augmented data for each iteration.
%We test this for all iterations - the regressor is trained for all community partitions generated.

For regression we use
%To summarize, we train one regressor per pair (target; iteration).
%The regressor models used are: 
linear regressor, ridge, LASSO and SVR (support vector regressor). To set the hyperparameters, we apply a grid search. We perform
%to the hyperparameters of these models is conducted with 
10-fold cross validation, and we evaluate with
%. The error metrics used are: 
mean absolute error (MAE), mean squared error (MSE) and $R^2$. 
% A grid search over the regressors is performed and an hyper-parameter tuning on each regressor is also executed. The pair regressor/hyperparameter that provides the highest $R^2$ is selected and only the results from the best performing pair are reported. 

By using the post-treatment score (at $t_1$) of each layer as target and the communities at each iteration as input, we also assess to what extent we can predict a post-treatment score for some layer without exploiting the pre-treatment data (at $t_0$) of this layer. In particular, assume that the post-treatment score of questionnaire/layer $l$ is used as target and that we use as input the communities learned over the layers $l_1$ (first iteration) and $l_2$ (second iteration), where $l\not\in\{l_1,l_2\}$. If the prediction quality is high in the evaluation setting for a given phenotype, then we expect that we can predict the score of $l$ at $t_1$ without recording layer $l$ for this phenotype at all.

% \textcolor{gray}{Besides, we test it across all questionnaires (target variables). For example, in iteration 1 we only have a single-layer network but since the nodes represent the patients, we may have information about their scores in another questionnaires. Therefore, we can train a regressor of the partition from iteration 1 using as the target variable of our regressor another questionnaire and study how transferable are the communities found with the one questionnaire.}

Lastly, we compare the contribution to predictive quality achieved by the communities found by \COBALT{} to the predictive quality achieved when using clusters built
%our algorithm 
with traditional clustering algorithms. The clustering techniques used are: (i) agglomerative hierarchical clustering \cite{agglo}, which we denote as `AHS'; (ii) BIRCH \cite{birch}, (iii) Gaussian Mixture Models with the Expectation Maximization algorithm \cite{gmm1, gmm2}, which we denote as `GMM\_EM'; (iv) HDBSCAN \cite{hdbscan}, (v) k-means \cite{kmeans} and (vi) OPTICS \cite{optics}. 
%The algorithms are not directly comparable, but the type of evaluation aforementioned makes it possible to compare them. The same logic is employed: a regressor is trained per clustering solution and the prediction errors of post-treatment data are compared.
It must be stressed that these clustering models serve only as baselines, because they operate under different conditions than \COBALT{}: they are cost-insensitive, since they exploit
%take
\textit{all} questionnaires/layers,
%scores as features, 
while \COBALT{}
%our community detection algorithm only 
uses only the layers up to a given iteration. Moreover,
%the data from all features when it gets to the last iteration. Hence, this should be taken into account when interpreting the results.
%
%Another contrast between clustering models and our proposed algorithm is that
these clustering algorithms do not handle missing values, hence they are trained only on patients that answered all the questionnaires.
%data without missing values are used to train them. 

%\subsection{Phenotype contribution, as affected by missingness}
%\subsection{The impact of missing nodes on the discovery of communities}
\subsection{Measuring how missingness affects phenotype quality}
\label{subs_impact_mv}
\COBALT{} is designed to deal with layers that contain only few nodes, i.e. layers for which only few entities (in our application: patients) have delivered data. To measure the influence of missingness, we gradually introduce missingness into a dataset that originally has no missing data. In this dataset,
%
%The ratio of missing values in the original dataset varies per feature. In order to ensure that we study the impact of missing values over all features ``fairly", we filter the original data so that only known data is kept for this experiment. Consequently, 
when removing an entity,
%we have $1-x\%$ of data per feature. When we delete a node, 
we remove its corresponding node from all layers.
%This signifies that we remove the whole data of the patient represented by that node.

The complete workflow is as follows. If the dataset in use contains missing values, we extract from it the subset $D$ that has no missing values. In the next step, we specify the maximum and minimum of the `missingness ratio', which we define the percentage of entities to be randomly selected and eliminated from $D$. Next, we perform a
%
%To assess the robustness of \COBALT{}, we investigate how missing nodes affect phenotype quality.
%%the final community structure results. Our strategy is to hide a certain number of nodes and test their impact on the model. This is accomplished by removing \textcolor{MyraFGcolor}{a randomly selected} subset of nodes from the total. 
%This translates into one hyper-parameter to tune: the ratio of missing nodes. A 
grid search between the minimum and maximum percentage and derive the corresponding subset of $D$ for each value in the grid; for our experiment, we vary the missingness
%is performed where the missing node 
ratio from $0.1$($10\%$ of entities removed) to $0.9$($90\%$ of the entities removed), with a step of $0.1$, i.e. of 10\%.
%In each run, we add $10\%$ more of missing nodes until we reach a total of $90\%$.
Finally, we run \COBALT{} on each derived dataset and measure quality as (i) modularity at each iteration and (ii) predictive quality for the iteration with the best modularity.

%To conduct a sensitivity analysis of the impact of missing data in our model, we examine the modularity of the partitions with different missing data ratios, as well as the performance of the prediction of post-treatment data, using the community ID, age, gender and the pre-treatment data as the features of the predictor. This is explained in more detail in \ref{subsection_evaluation}.
%
%\COBALT{} is designed to deal with layers that contain only few nodes, i.e. layers for which only few patients have delivered data. To assess the effects of missingness on predictive quality, we fit a regressor per parition and assess which partition, per missing node ratio, best predicts the target variable. Due to the existence of few data points for training, specially after removing nodes, it is only feasible to perform this evaluation for ratios of 10\% and 20\%.

\section{Results and Discussion}
\label{results}
For our evaluation, we apply \COBALT{} on the tinnitus patient dataset described in section \ref{subsec:uhregData}. This dataset contains 5 questionnaires: in a realistic scenario, the addition of layers would have stopped at an upper boundary of cost/budget; for the evaluation, we add all layers and study the effects of adding each one. \COBALT{} adds the layers in the following order: (1) THI, (2) MDI, (3) TQ, (4) TBF12 and (5) TFI questionnaire. This indicates that the benefit of adding THI is the highest and the benefit of adding TFI is low.

%In subsection \ref{search_alg_results}, we report on the performance of \COBALT{} for phenotype construction, in comparison to 
%%show the results from the proposed search algorithm explained in subsection \ref{search_algorithm}. These results are also compared with 
%traditional clustering algorithms. In subsection \ref{missingness_results} we elaborate on
%%display the results from our experiments to study
%the impact of missing nodes 
%%in our search algorithm.
%on \COBALT{}.

%\subsection{Search algorithm}
%\subsection{Phenotype quality vs layer cost}
\subsection{Modularity vs cost per iteration}
\label{search_alg_results}

\begin{figure}[htb]
\centering
   \includegraphics[height=5.3cm,keepaspectratio]{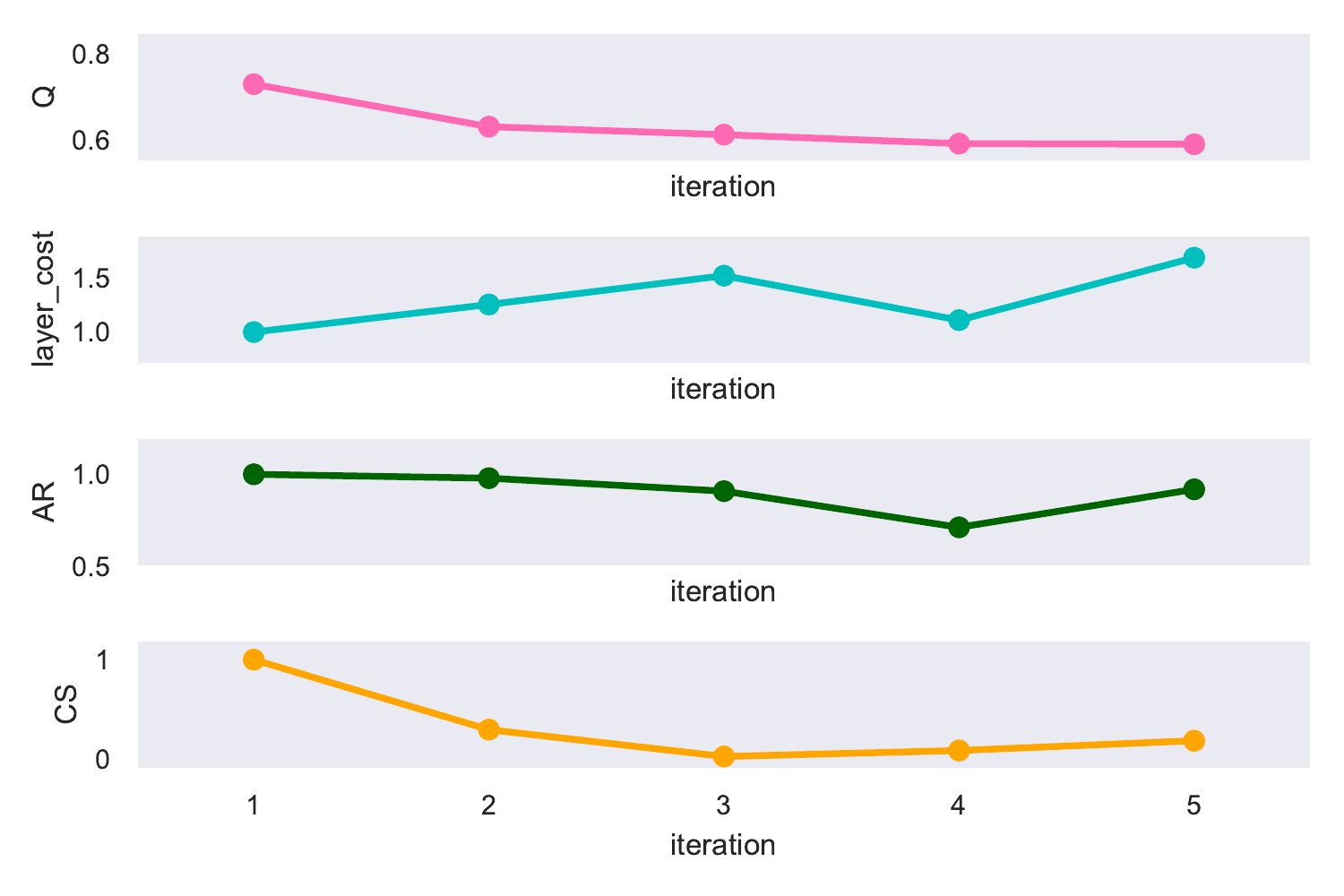}
\caption{Evolution of modularity (uppermost subfigure), cost (middle upper subfigure), availability ratio (middle lower subfigure) and community similarity (lowermost subfigure), computed as layers are added by \COBALT{}, one at a time.}
\label{fig:mod_layer_cost}
\end{figure}

Figure \ref{fig:mod_layer_cost} shows the evolution of the modularity $Q$ per iteration, as well as the layer cost and availability ratio of the selected layer at that iteration.
The modularity decreases as layers are added (uppermost subfigure), while the community similarity stagnates at 0 (lowermost subfigure), indicating that there is substantial difference among the communities built while adding layers. Despite the decrease in modularity, it is important to note that the drop is from $0.730$ to $0.591$, indicating that
%. Even at iteration 5, the modularity can be considered a high value and may indicate that 
there are underlying structures in the MLN. The visualizations presented next deliver insights on how communities change as layers are added.

At the same time, the cost of adding a layer is increasing for all but the 4th layer (cf. middle upper subfigure). This increase is mostly due to the decrease in the availability ratio (cf. middle lower subfigure), since community similarity is always low. The availability ratio decreases slowly,
%(the lowermost position at the vertical axis is much higher than 0.5), so that 
hence \textit{SC1} can be triggered already at the 3rd or at the 4th iteration.

%The cost increase is explained by the drop in the availability ratio (see Table \ref{tab:materials_table}).%\textcolor{red}{Table with the number of patients per layer must be cited here}. 
%The cost decrease at the 4th layer, which corresponds to the small questionnaire TBF12, indicates that the availability ratio does not change much between the previous iteration and this one, and/or that the communities built when adding this layer are rather different from those built thus far.
%There is no clear relation between the cost of the layer added. Regardless of the layer cost being high or low, the modularity drops with additional layers. 

\subsection{Community visualization}

In Figures \ref{fig:thi_iteration_1}
%, \ref{fig:iteration_2}, \ref{fig:iteration_3}, \ref{fig:iteration_4} and
to \ref{fig:iteration_5} we show the communities as \COBALT{} adds layers. The visualization is two-dimensional and therefore we show one subfigure per layer.
%from the 2nd iteration onwards.

%display the visualization of the networks correspondent to iterations 1, 2, 3, 4 and 5, respectively. This illustration is 2D and therefore each network represents only one layer. The colors of the nodes denote the community to which they belong to. The blank areas are empty spaces where there are no edges neither nodes. The layout used for visualization was the Fruchterman-Reingold layout \cite{Fruchterman1991}. Nodes that are positioned closer to one another have stronger connections between them than with the ones located farther apart. 

\subsubsection{Visualization of one layer}
\begin{figure}[htb]
\centering
       \includegraphics[height=5cm,keepaspectratio]{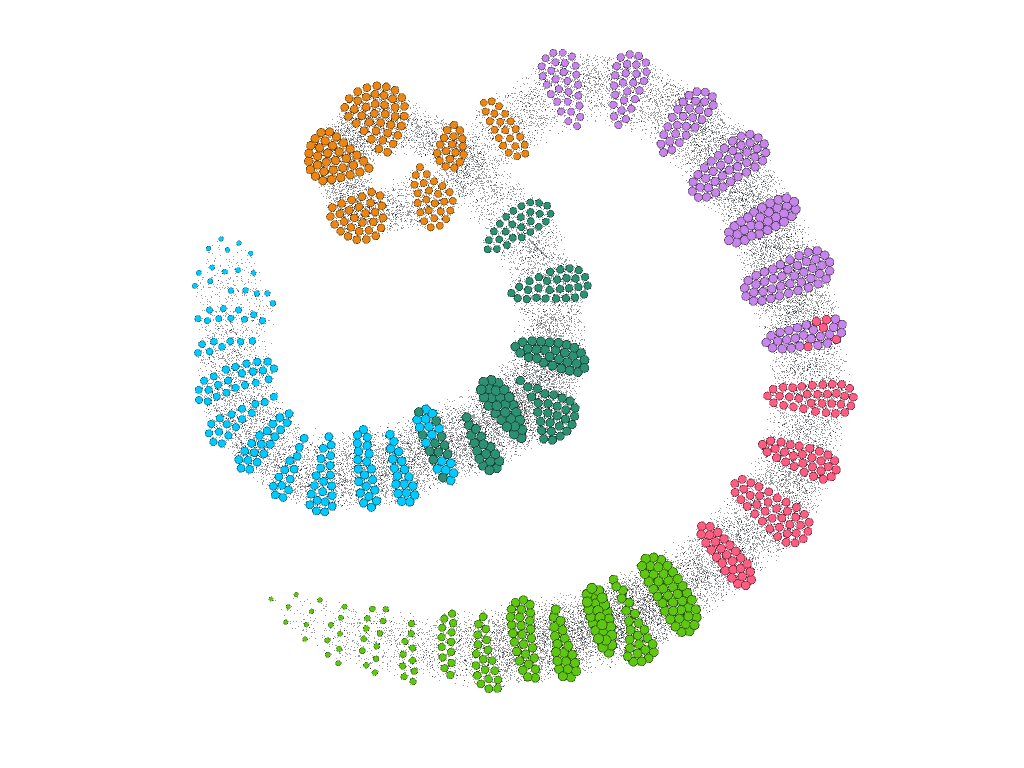}          
\caption{1st iteration: 6 communities on the THI network; $Q=0.730$. Different colors represent different communities.}
\label{fig:thi_iteration_1}
\end{figure}

In Figure \ref{fig:thi_iteration_1} we depict the 6 communities 
%found by the Leiden algorithm
in the THI layer, which is chosen by \COBALT{} in the first iteration.
%the Leiden algorithm found 6 communities in the single-layer network that represents the THI questionnaire data. 

The communities mark 
%separate the graph in 
areas of a single color. There is one subgraph with nodes of the `blue' and the `darkgreen' communities, and one subgraph with both `purple' and `pink' nodes, but these two subgraphs are only small parts of the network.
%in comparison to the graph as a whole.
Hence, the visualization scheme captures properly community separation 
%of the network into communities 
and clearly marks the areas with nodes from more than one community.

%The partition seems to agree with the visualization algorithm, i.e. nodes that are located in a certain area belong to the same community. There are, however, two exceptions which concern the blue and green community and the purple and pink communities. There are areas of the graph where there is a mix between green and blue nodes and purple and pink nodes.

\begin{figure*}[tb]
  \includegraphics[width=16cm,keepaspectratio]{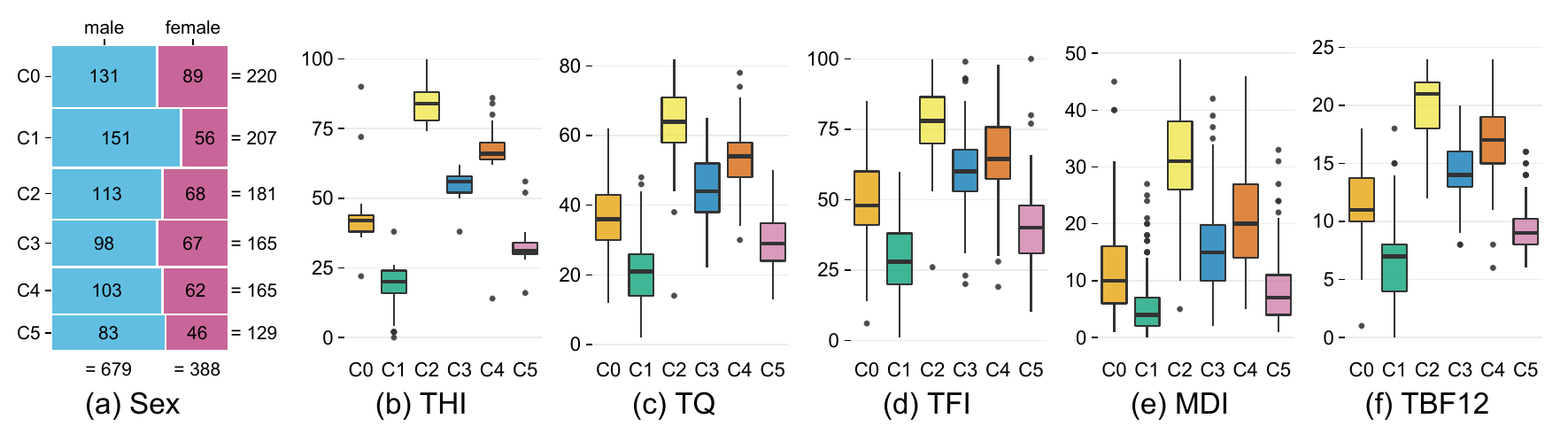}
  \caption{The 6 communities of the THI network added by \COBALT{} as 1st layer, depicting the value distributions inside each questionnaire.
  %\textcolor{anyFGcolor}{CLARA, since we use only the 1st layer, we have missing values in the MDI, TQ etc, right? Can we say how many people are in the MDI part, in the TBF12 part etc? Is it possible to have the same ordering as in Table 4, i.e. THI first, then MDI, then TQ, then TFI, then TBF12?}
  }
  \label{fig:results_cd}
\end{figure*}

%\subsection{Communities in the first layer added by \COBALT{}}
\subparagraph{Score distributions}
Figure \ref{fig:results_cd} depicts the distribution of the questionnaire scores for each of the 6 communities of this 1st iteration of \COBALT{}.
%A visualization of the distribution of the questionnaire scores across the communities from the THI layer (iteration 1) %best partition is shown in Figure \ref{fig:results_cd}.

The first column of Figure \ref{fig:results_cd} shows the value distribution for sex, with one row per community. Subsequently, we show one column per questionnaire and, inside it, one boxplot per community. We see that all communities are rather homogeneous with respect to the THI questionnaire: the small boxes indicate small variance. The variance increases for the other questionnaires; this is expected, since \COBALT{} considered only the THI data to build these communities.

There are remarkable differences among some of the communities.
%The questionnaire score ranges differ per community: 
%for example, 
Community C1 (green boxes for the questionnaires) has the lowest average value in each questionnaire, while
%Another example is 
community C2 (fade-yellow boxes) has the highest average values, indicating
%for all questionnaire scores. This could imply that 
C1 and C2 have considerably different characteristics.
%Communities C3 (blue boxes) and C4 (dark-orange boxes) have rather high average values for all questionnaires except MDI, with the averages of C3 being a bit lower and the variance a bit smaller than the corresponding C4 values.
In the context of the data used, it seems
%this might mean 
that C1 accommodates the patients with the most mild tinnitus symptoms, while C2 accommodates the patients with the most severe tinnitus symptoms.

\subsubsection{Communities across layers}
\begin{figure}[h!]
\begin{subfigure}{.22\textwidth}
  \centering
  % include first image
  \includegraphics[width=.9\linewidth,trim=0 0 120 0]{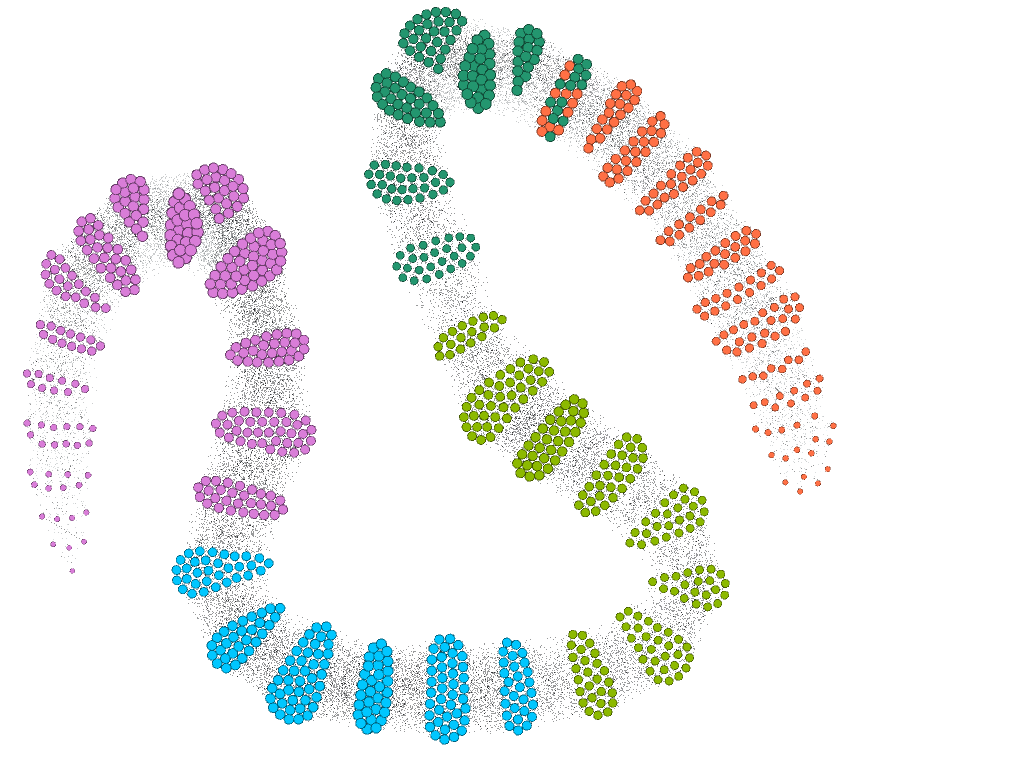}  
  \caption{THI network}
  \label{fig:thi_iteration_2}
\end{subfigure}
\begin{subfigure}{.22\textwidth}
  \centering
  % include second image
  \includegraphics[width=.9\linewidth,trim=100 0 100 100]{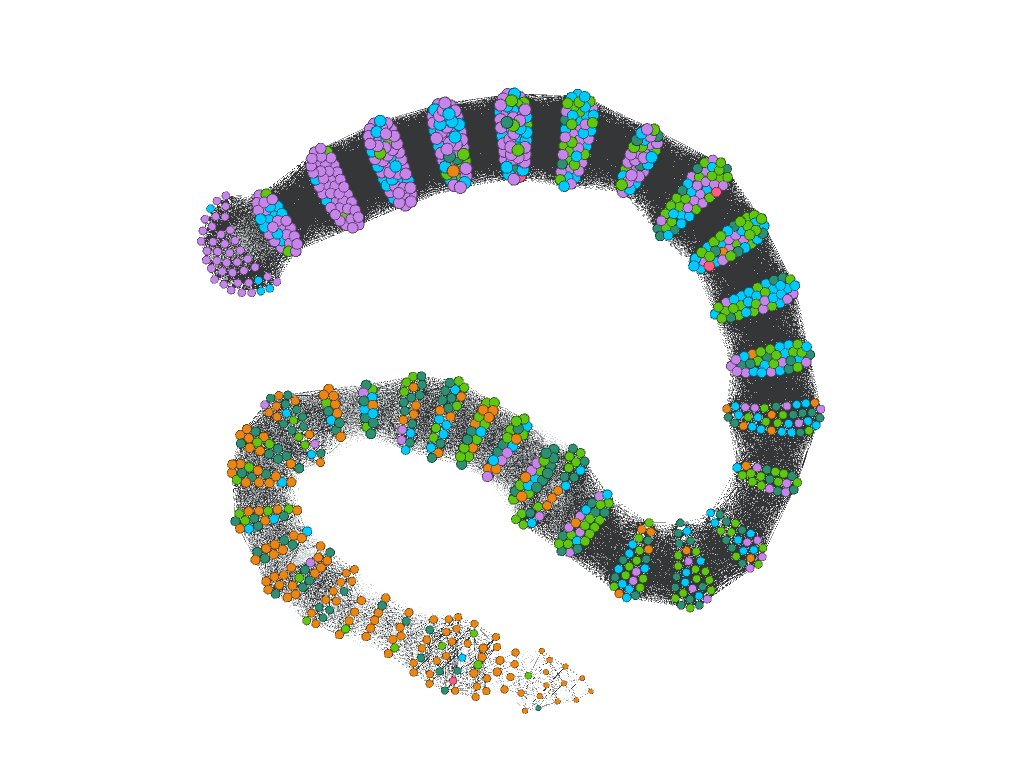}  
  \caption{MDI network}
  \label{fig:mdi_iteration_2}
\end{subfigure}
\caption{2nd iteration with MDI questionnaire as 2nd layer: 5 communities detected; $Q=0.632$.}
\label{fig:iteration_2}
\end{figure}

Figure \ref{fig:iteration_2} depicts the 5 communities found in the 2nd iteration, where \COBALT{} adds the MDI layer as the one with the lowermost cost.
%shows again the first layer selected (THI) plus the MDI layer. The Leiden algorithm detects the communities using data from both layers, leading to a different community structure when comparing with iteration 1. The MDI layer is the selected layer to be added in iteration 2, since it is the layer with the lowest cost.

As we have seen in the evolution curves of Figure \ref{fig:mod_layer_cost}, iteration 2 incurs a cost increase 
%(which would have stopped the iterative procedure of \COBALT{} in a real setting), that is also reflected in 
and a slight modularity drop. The visualization of the communities in the two layers makes this evident: there is a rather clear separation of colors in the THI network (cf. subfigure \ref{fig:thi_iteration_2}), but the colors in the MDI network (cf. subfigure \ref{fig:mdi_iteration_2}) are mixed. Hence, the use of the inter-layer similarities and the similarities inside the MDI network did not contribute to a good modularity score.
%to the addition of useless layers  than the modularity score.
%, which has increased but is still high.
%This indicates that the modularity score (which is still high) is due to the homogeneity of the communities inside the THI network.

%than by the intra-layer similarity inside the MDI network and the inter-layer similarity between the two networks.

%By including this new layer, the visualization of the MDI network (which only takes into account the data from this layer) locates the nodes from different communities close to each other. The nodes are visibly scattered in Figure \ref{fig:mdi_iteration_2}. This does not necessarily imply that the community assignments are poor. In fact, this is the expected behavior: the algorithm uses not only data from the MDI layer, but also data from the THI layer, making partitioning decisions relying on both values. Figure \ref{fig:thi_iteration_2} depicts the partition of this two-layer network with respect to the THI layer, which appears to split the nodes well with respect to the THI layer. The number of communities, on the other hand, has reduced to 5. As a result, the insertion of the MDI layer has altered the partition, but not much in relation to the THI layer.

\begin{figure}[htb]
\begin{subfigure}{.15\textwidth}
  \centering
  % include third image
  \includegraphics[width=.9\linewidth]{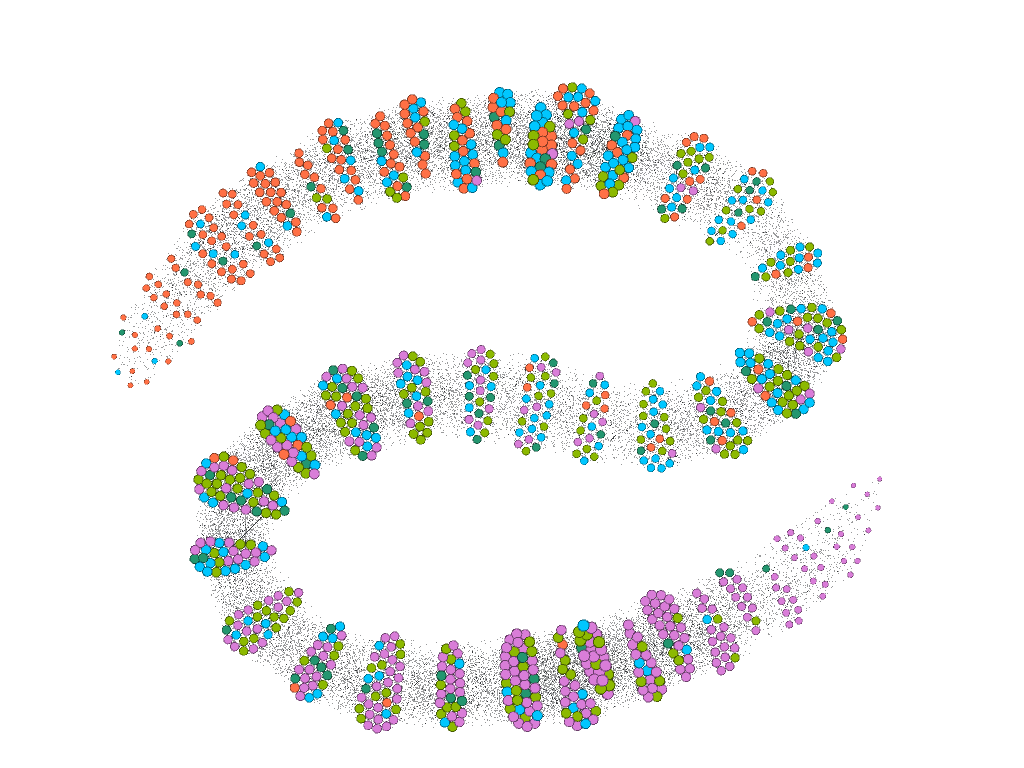}  
  \caption{THI network}
  \label{fig:thi_iteration_3}
\end{subfigure}
\begin{subfigure}{.15\textwidth}
  \centering
  % include fourth image
  \includegraphics[width=.9\linewidth]{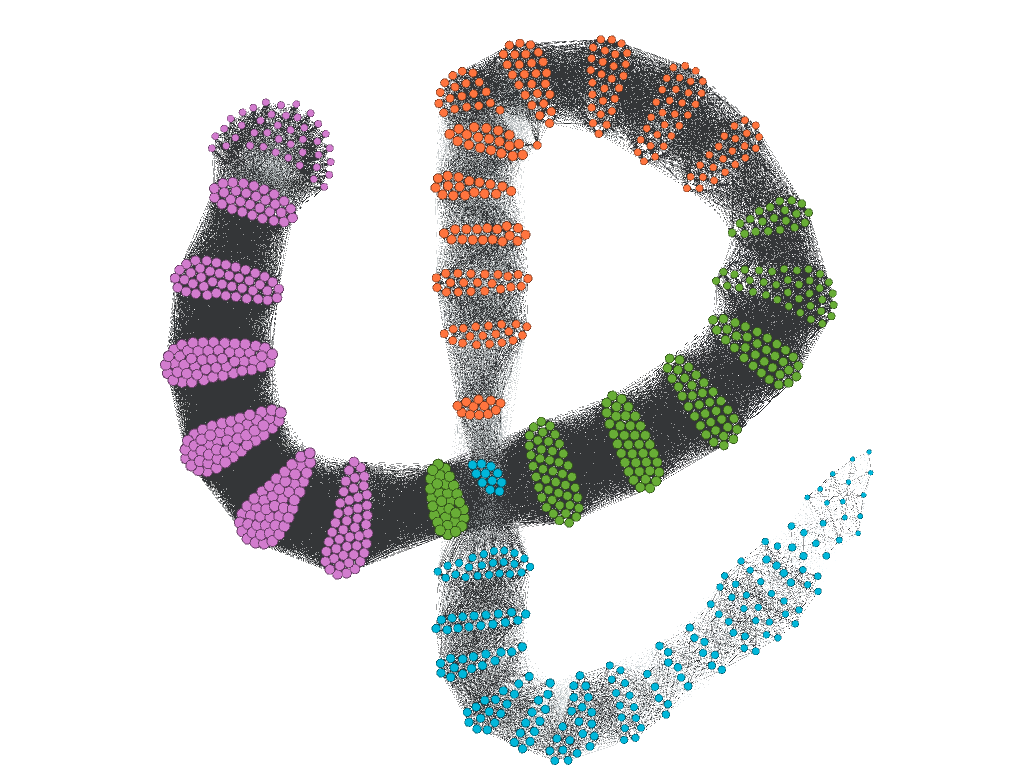}  
  \caption{MDI network}
  \label{fig:mdi_iteration_3}
\end{subfigure}
%\\
\begin{subfigure}{.15\textwidth}
  \centering
  % include fourth image
  \includegraphics[width=.9\linewidth,trim=30 50 80 30]{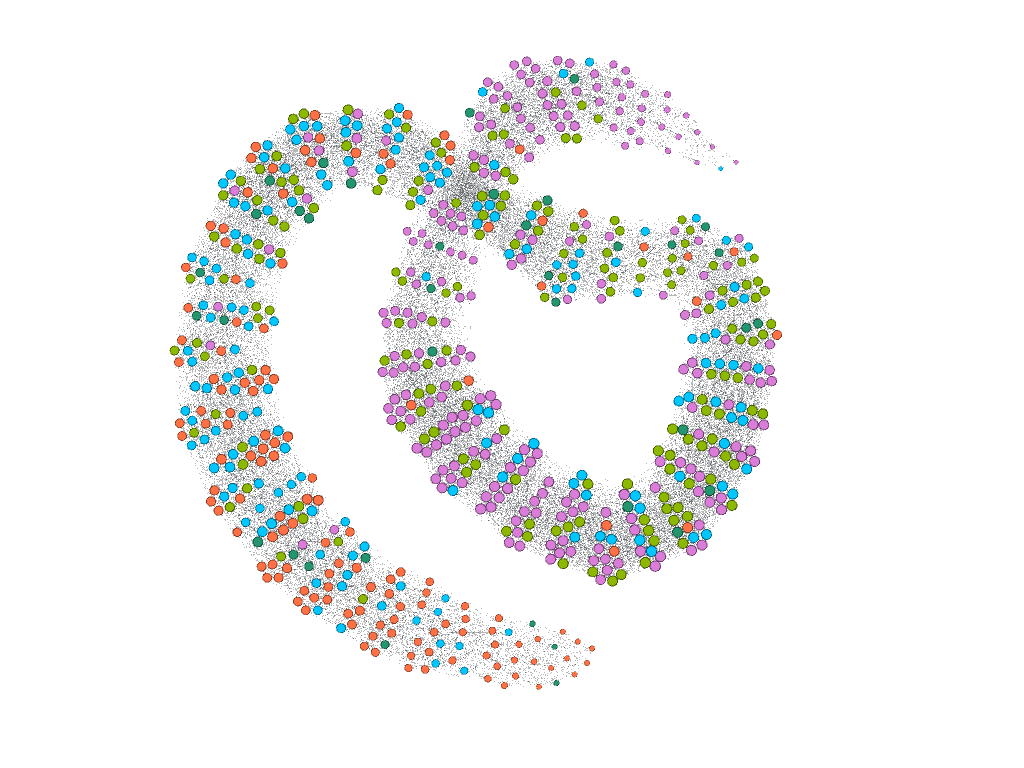}  
  \caption{TQ network }
  \label{fig:tq_iteration_3}
\end{subfigure}
\caption{3rd iteration with TQ questionnaire added as 3rd layer: 4 communities detected; $Q=0.613$.}
\label{fig:iteration_3}
\end{figure}

Figure \ref{fig:iteration_3} shows the 4 communities found in the 3rd iteration, where \COBALT{} added the TQ layer. From this iteration on, community induction is driven by the node similarities inside the MDI layer, leading to more homogeneous communities in the MDI layer, while the community colors in the other layers are mixed. 

\begin{figure}[htb]
\begin{subfigure}{.22\textwidth}
  \centering
  % include third image
  \includegraphics[width=.9\linewidth]{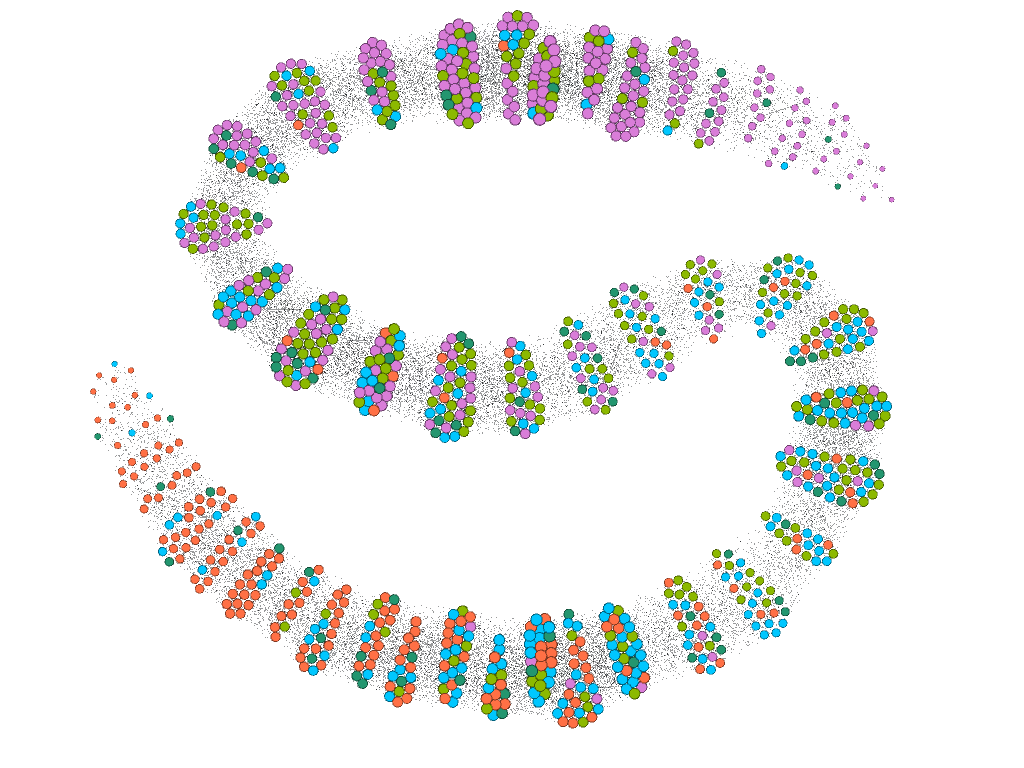}  
  \caption{THI network}
  \label{fig:thi_iteration_4}
\end{subfigure}
\begin{subfigure}{.22\textwidth}
  \centering
  % include fourth image
  \includegraphics[width=.9\linewidth]{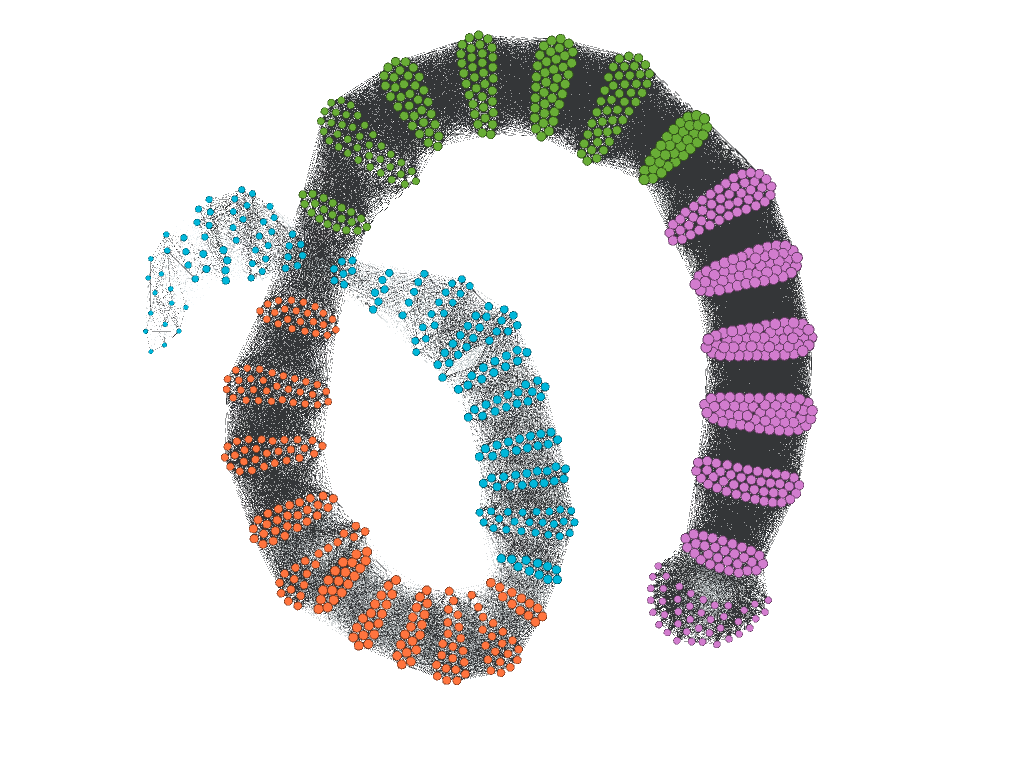}  
  \caption{MDI network}
  \label{fig:mdi_iteration_4}
\end{subfigure}
%\\
\begin{subfigure}{.22\textwidth}
  \centering
  % include fourth image
  \includegraphics[width=.9\linewidth]{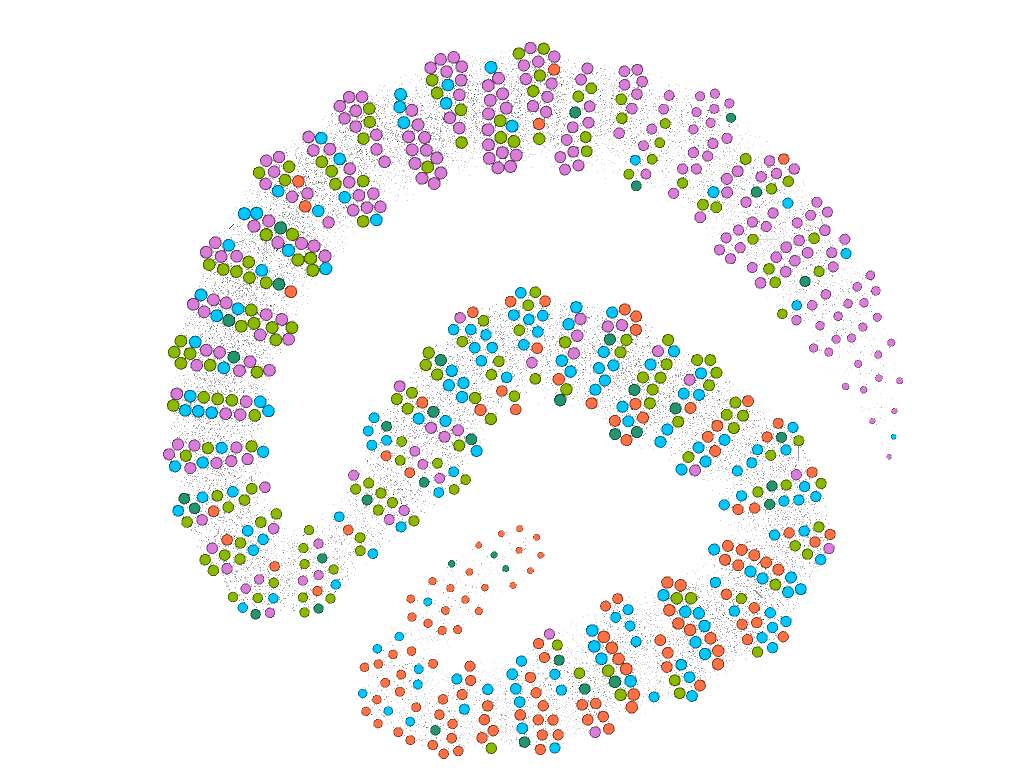}  
  \caption{TQ network}
  \label{fig:tq_iteration_4}
\end{subfigure}
\begin{subfigure}{.22\textwidth}
  \centering
  % include fourth image
  \includegraphics[width=.9\linewidth,trim=0 0 0 0]{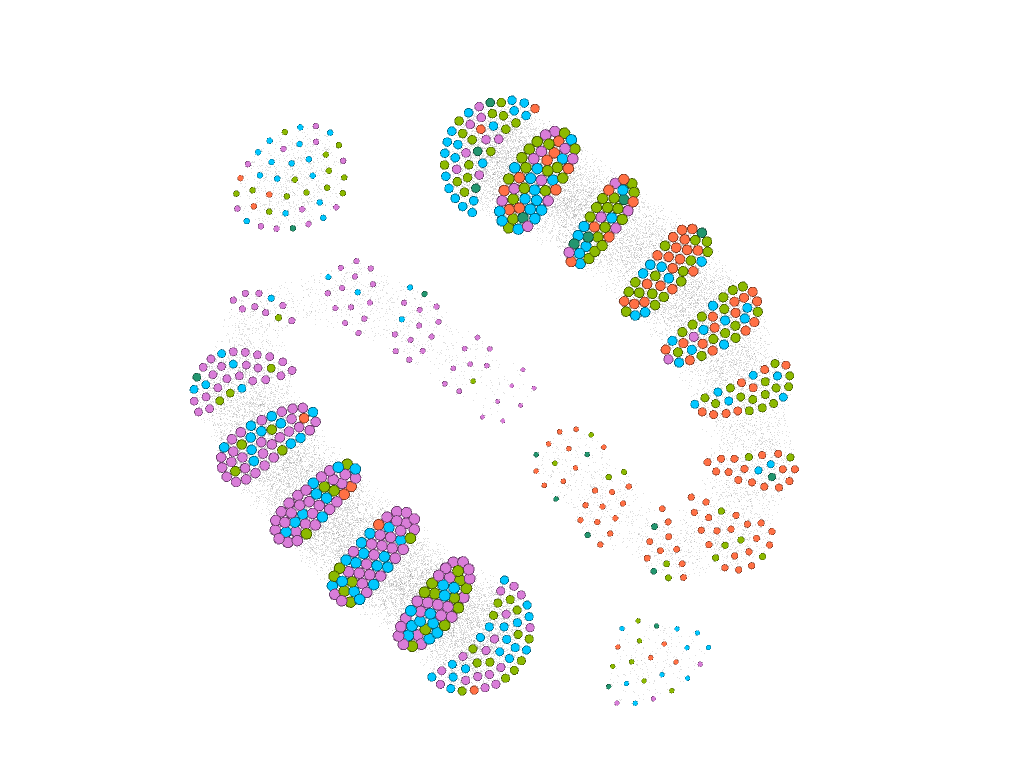}  
  \caption{TBF12 network}
  \label{fig:tbf12_iteration_4}
\end{subfigure}
\caption{4th iteration with TBF12 added as 4th layer: 4 communities detected; $Q=0.592$.}
\label{fig:iteration_4}
\end{figure}

A further remarkable aspect in Figure \ref{fig:iteration_3} is the high density of the MDI network: the patients are very similar to each other inside this layer. This might have led to communities that are not clearly separated.
%so that the graph pruning step (cf. subsection \ref{methodology_pruning}) had a smaller effect when compared with other layers.
%In this partition, nodes seem to be well separated with respect to the MDI layer, whereas in the other two layers the colors appear dispersed in many areas of the networks.
This can also be seen in Figure \ref{fig:iteration_4}, which depicts the 4 communities found when \COBALT{} adds TBF12 as 4th layer.

As in the 3rd iteration,
%In this iteration, once again 
nodes are well-separated into communities with respect to the MDI layer, but not with respect to the other layers. The number of communities is also the same as before: although this does not imply that the communities are exactly the same, it indicates
%remains constant (4 communities). This suggests that 
adding the TBF12 layer 
%had little effect on the structure of the communities detected. 
%%Nonetheless, this does not mean that the communities are exactly the same as in iteration 3. 
%Our main motivation when employing a layer-cost based model is precisely to leave to the final iterations the layers that offer the least additional information. The results of the 4th iteration suggest that indeed adding layer TBF12 
does not add much
%a high amount of relevant 
information to the previous MLN.
%of iteration 3. 

\begin{figure}[htb]
\begin{subfigure}{.19\textwidth}
  \centering
  % include third image
  \includegraphics[width=.98\linewidth]{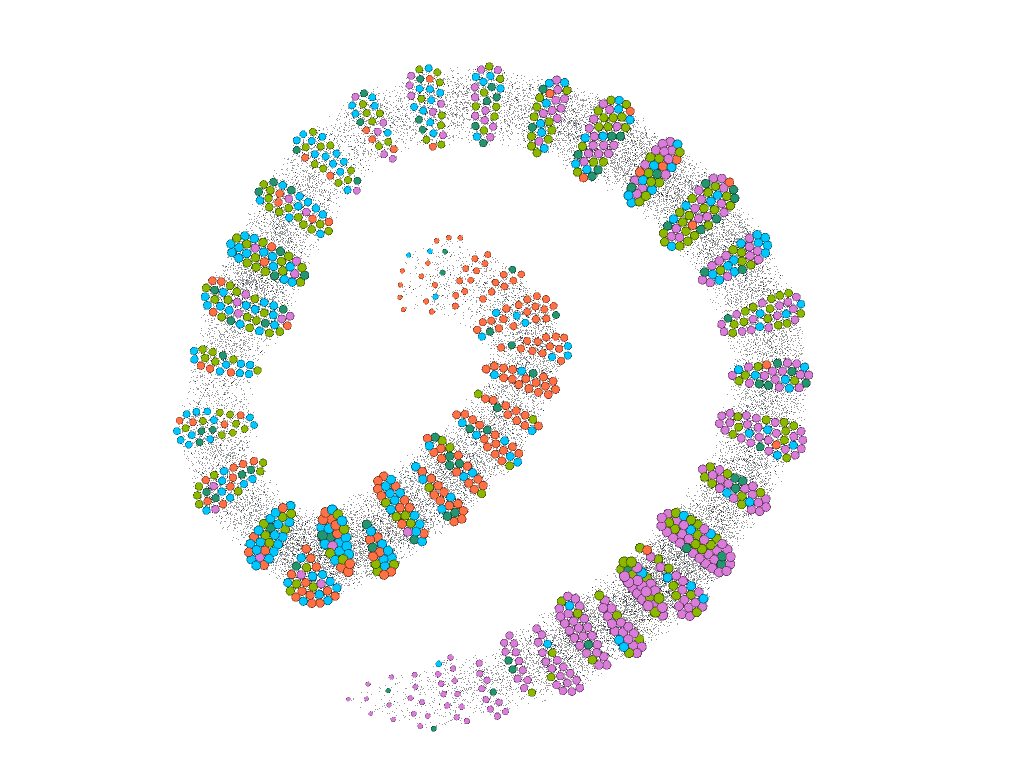}  
  \caption{THI network}
  \label{fig:thi_iteration_5}
\end{subfigure}
\begin{subfigure}{.19\textwidth}
  \centering
  % include fourth image
  \includegraphics[width=.98\linewidth]{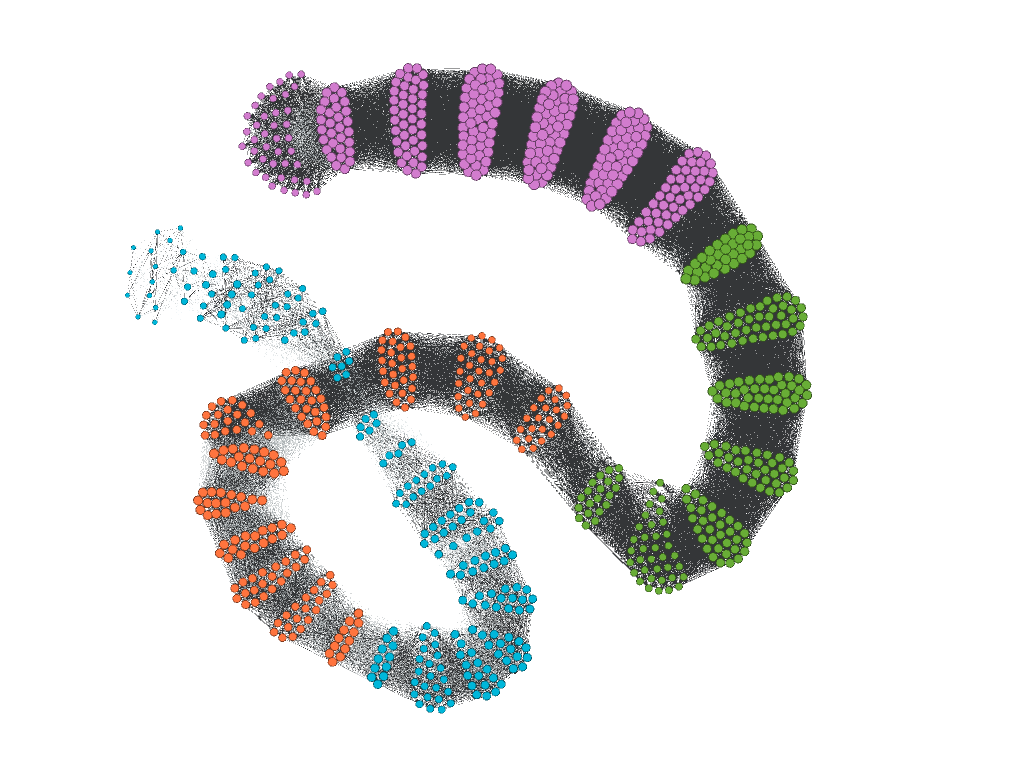}  
  \caption{MDI network}
  \label{fig:mdi_iteration_5}
\end{subfigure}
\begin{subfigure}{.19\textwidth}
  \centering
  % include fourth image
  \includegraphics[width=.98\linewidth]{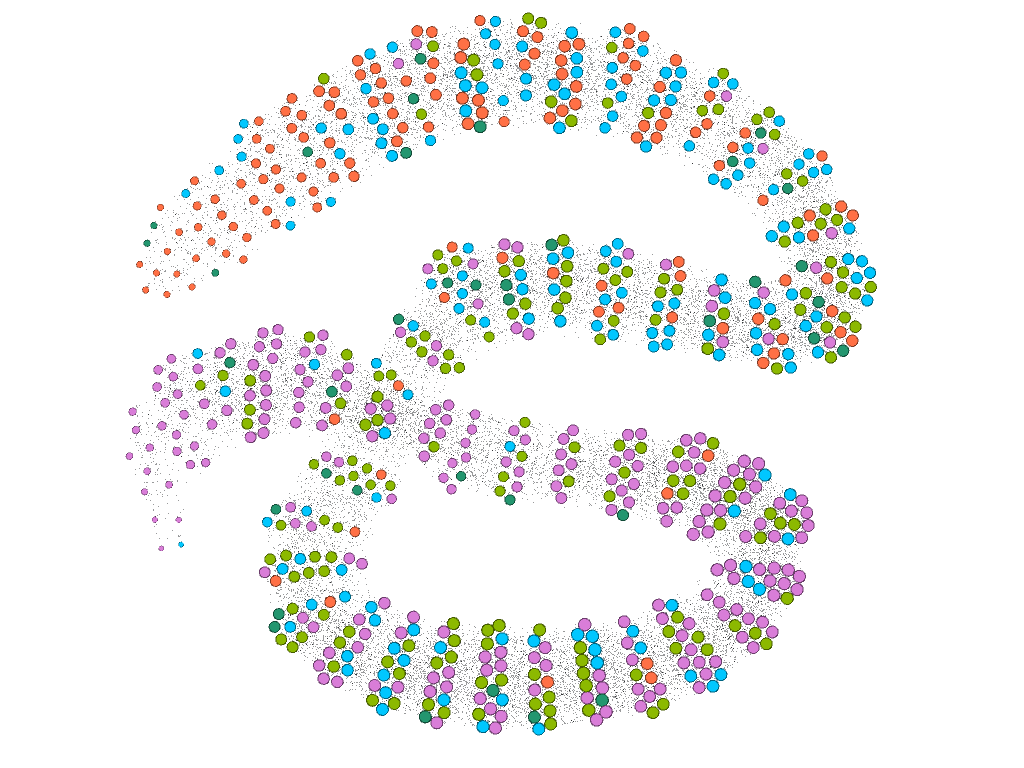}  
  \caption{TQ network }
  \label{fig:tq_iteration_5}
\end{subfigure}
\begin{subfigure}{.19\textwidth}
  \centering
  % include fourth image
  \includegraphics[width=.98\linewidth,trim=-10 0 0 0]{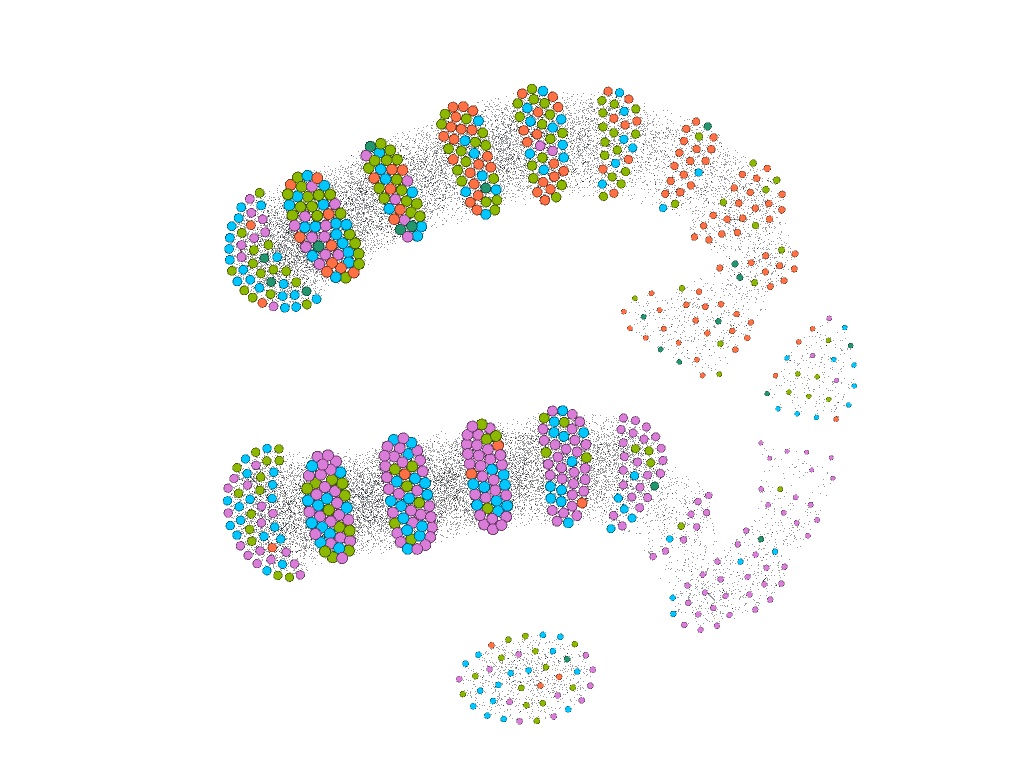}  
  \caption{TBF12 network }
  \label{fig:tbf12_iteration_5}
\end{subfigure} 
\begin{subfigure}{.19\textwidth}
  \centering
  % include fourth image
  \includegraphics[width=.98\linewidth,trim=0 0 0 0]{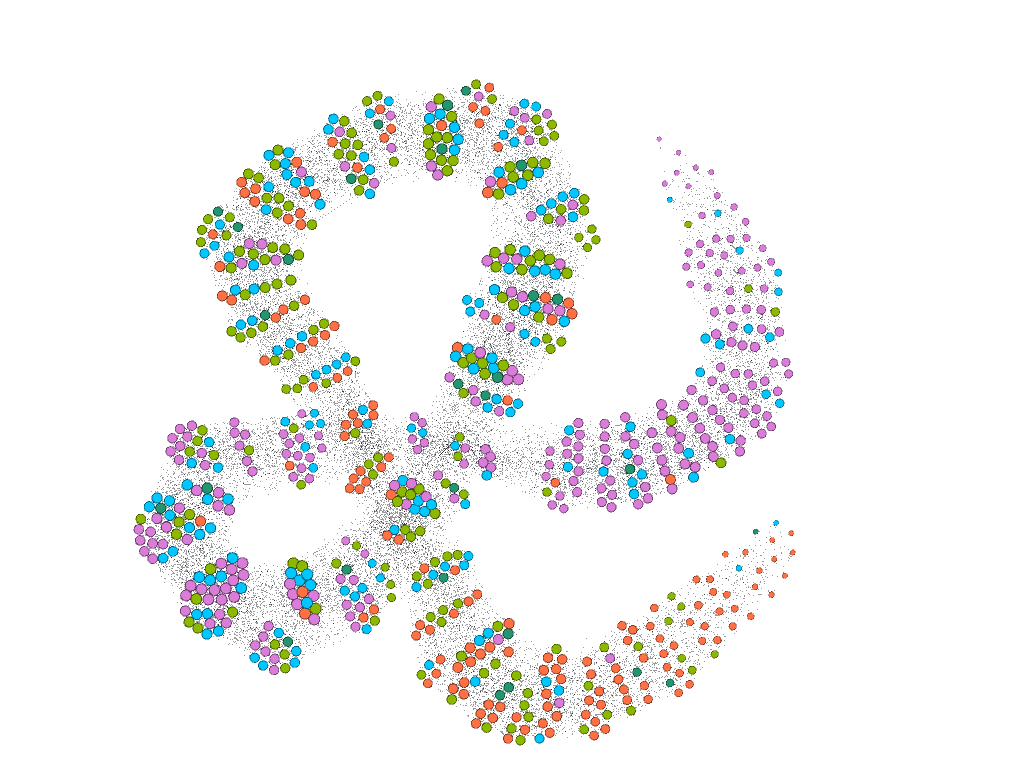}  
  \caption{TFI network}
  \label{fig:tfi_iteration_5}
\end{subfigure}
\caption{5th iteration, last layer (TFI) added: 4 communities detected; $Q=0.591$.}
\label{fig:iteration_5}
\end{figure}

Finally, Figure \ref{fig:iteration_5} shows the 4 communities found in the 5th iteration, when the TFI layer is added as last one. 

As before, the communities are well separated in the MDI network, but not in the other networks.
%displays the communities discovered using the 5 layers. The last layer to be added, and hence the more costly, is the TFI layer. The algorithm found again 4 communities and the nodes are also well separated with respect to the MDI network.
The modularity 
%of this partition 
($0.591$) is also very close to the modularity value of the 4th iteration ($0.592$).

Summarizing, the visualization scheme shows a deterioration of community quality as layers are added. The evolution of layer cost and of its two factors capture this deterioration well; the
%better than the modularity, which decreases only slowly. The 
\textit{SC1} stopping criterion would have stopped the MLN expansion after the 3rd iteration by the latest.

In our application scenario, communities serve as phenotypes, intended to contribute to prediction. For that reason, we next report on how the communities contributed to predictive quality.

%\subsection{Contribution of the phenotypes found by \COBALT{} to outcome prediction}

\subsection{Phenotypes for prediction}
%The predictive task we investigate is the assessment of the post-treatment scores of the patients, i.e. at $t_1$, using following features: age, gender, questionnaire score before the treatment, i.e. at $t_0$, as well as the communities returned by \COBALT{}, again at $t_0$. 
%We denote the post-treatment moment as $t_1$ and the baseline as $t_0$. 
According to our evaluation design, we compare the prediction quality achieved when using \COBALT{} communities to that achieved by the Baseline regressors and to the quality achieved when using clustering instead of \COBALT{}.

The results are on Table \ref{tab:predictions_best}, which we discuss in detail hereafter. For each of the 5 scores at $t_1$, we depict the performance as MSE and MAE (where smaller values are better) and $R^2$ (where larger values are better). Values marked in \textbf{boldface} are the best achieved for predicting a questionnaire score.

\begin{table*}[tb]
\resizebox{\textwidth}{!}{%
\begin{tabular}{llllllllllllllllll}
\cline{2-18} 
\multicolumn{1}{c}{} & \multicolumn{2}{c}{\textbf{Partition quality}}             & \multicolumn{3}{c}{\textbf{THI score at $t_1$}}                                            & \multicolumn{3}{c}{\textbf{MDI score at $t_1$}}                                            & \multicolumn{3}{c}{\textbf{TQ score at $t_1$}}                                             & \multicolumn{3}{c}{\textbf{TBF12 score at $t_1$}}                                            & \multicolumn{3}{c}{\textbf{TFI score at $t_1$}}                                          \\ \cmidrule(lr){2-3} \cmidrule(lr){4-6} \cmidrule(lr){7-9} \cmidrule(lr){10-12} \cmidrule(lr){13-15} \cmidrule(lr){16-18} 
\multicolumn{1}{c}{\textbf{Model}}               & \multicolumn{1}{c}{silhouette} & \multicolumn{1}{c}{Q}     & \multicolumn{1}{c}{MSE} & \multicolumn{1}{c}{MAE} & \multicolumn{1}{c}{$R^2$} & \multicolumn{1}{c}{MSE} & \multicolumn{1}{c}{MAE} & \multicolumn{1}{c}{$R^2$} & \multicolumn{1}{c}{MSE} & \multicolumn{1}{c}{MAE} & \multicolumn{1}{c}{$R^2$} & \multicolumn{1}{c}{MSE} & \multicolumn{1}{c}{MAE} & \multicolumn{1}{c}{$R^2$} & \multicolumn{1}{c}{MSE} & \multicolumn{1}{c}{MAE} & \multicolumn{1}{c}{$R^2$} 

\\     \cmidrule(lr){2-3} \cmidrule(lr){4-6} \cmidrule(lr){7-9} \cmidrule(lr){10-12} \cmidrule(lr){13-15} \cmidrule(lr){16-18}

\textbf{Baseline}                           &                                &                           &  $13.1$                        &             $287.6$            &           $0.569$                               &      $5.1$                   &       $43.0$                  &       $0.686$                                   &              $9.9$           &           $159.9$              &               $0.394$             &              $3.1$           &       $13.2$                  &  $0.182$               &                $10.2$         &             $158.4$            &               $0.637$                           
\\
\cline{2-18} 

\textbf{\COBALT{}}                                 & \multicolumn{1}{c}{}           & \multicolumn{1}{c}{}      & \multicolumn{1}{c}{}    & \multicolumn{1}{c}{}    & \multicolumn{1}{c}{}                     & \multicolumn{1}{c}{}    & \multicolumn{1}{c}{}    & \multicolumn{1}{c}{}                     & \multicolumn{1}{c}{}    & \multicolumn{1}{c}{}    & \multicolumn{1}{c}{}                     & \multicolumn{1}{c}{}    & \multicolumn{1}{c}{}    & \multicolumn{1}{c}{}                     & \multicolumn{1}{c}{}    & \multicolumn{1}{c}{}    & \multicolumn{1}{c}{}                     \\
\hspace{7pt}iteration 1                        & \multicolumn{1}{c}{}           & \multicolumn{1}{c}{\underline{0.730}} & \textrm{9.3}                     & \textrm{130.9}                  & \textrm{0.720}                                    & \textrm{4.9}                     & \textrm{43.0}                    & \textrm{0.707}                                    & $11.3$                    & $280.3$                   & $0.284$             & $2.8$     & $13.2$        & $0.613$                          & \textrm{11.1}                    & $346.4$                   & $0.344$                                   \\

\hspace{7pt}iteration 2                        & \multicolumn{1}{c}{}           & \multicolumn{1}{c}{0.632} & $12.3$                    & $228.2$                   & $0.439$                                    & $5.4$                     & $62.6$                    & $0.532$                                    & $9.6$                     & $143.4$                   & $0.521$                & $2.4$                     & $9.8$                     & $0.798$                     & \textbf{\underline{10.1}}           & \textbf{\underline{148.2}}          & \textbf{\underline{0.730}}                                                              \\

\hspace{7pt}iteration 3                        & \multicolumn{1}{c}{}           & \multicolumn{1}{c}{0.613} & $9.6$                     & $154.6$                   & $0.563$                                    & $6.2$                     & $75.4$                    & $0.227$                                    & $9.2$                     & $123.7$                   & $0.573$                       & \textbf{\underline{2.1}}  & \textbf{\underline{5.9}}  & \textbf{\underline{0.866}}                 & $13.1$                    & $294.2$                   & $0.354$                                                           \\

\hspace{7pt}iteration 4                        & \multicolumn{1}{c}{}           & \multicolumn{1}{c}{$0.592$} & $13.0$                    & $265.8$                   & $0.606$                                    & \underline{4.4}            & \underline{36.9}           & \textbf{\underline{0.715}}                           & \underline{$7.6$}                     & \underline{$93.9$}                    & \underline{$0.622$}                  & $2.5$                     & $8.7$                     & $0.738$                        & $11.2$                    & $177.9$                   & $0.621$                                                                  \\

\hspace{7pt}iteration 5                        &                                &  \multicolumn{1}{c}{$0.591$}                     & \underline{$8.6$}                     & \underline{$103.5$}                   & \underline{$0.748$}                                    & $5.9$                     & $68.7$                    & $0.302$                                    & $11.6$                    & $227.1$                   & $0.379$                   & $3.2$                     & $13.4$                    & $0.595$                                                     & $14.9$                    & $356.8$                   & $0.359$                                    \\

\cline{2-18} 
\textbf{Clusterers}                                & \multicolumn{1}{c}{}           & \multicolumn{1}{c}{}      & \multicolumn{1}{c}{}    & \multicolumn{1}{c}{}    & \multicolumn{1}{c}{}                     & \multicolumn{1}{c}{}    & \multicolumn{1}{c}{}    & \multicolumn{1}{c}{}                     & \multicolumn{1}{c}{}    & \multicolumn{1}{c}{}    & \multicolumn{1}{c}{}                     & \multicolumn{1}{c}{}    & \multicolumn{1}{c}{}    & \multicolumn{1}{c}{}                     & \multicolumn{1}{c}{}    & \multicolumn{1}{c}{}    & \multicolumn{1}{c}{}                     \\
\hspace{7pt}AHC                                & $0.479$                          &                           & $9.9$                     & $198.9$                   & $0.565$                                    & $8.8$                     & $116.8$                   & $0.203$                                    & $8.2$                     & $110.2$                   & $0.355$            & $2.1$                     & $6.4$                     & $0.765$ 
& $14.8$                    & $389.3$                   & $-0.085$                                                      

\\
\hspace{7pt}BIRCH                              & $0.482$                          &                           & $10.1$                    & $183.1$                   & $0.576$                                    & $5.3$                     & $53.6$                    & $0.599$                                    & $6.7$                     & $61.5$                    & $0.455$                   & $3.8$                     & $15.7$                    & $0.671$                  & $11.7$                    & $325.6$                   & $0.108$                                                                       \\

\hspace{7pt}GMM\_EM                                 & $0.349$                          &                           & $8.4$                     & $138.5$                   & $0.564$                                    & $6.8$                     & $82.1$                    & $0.545$                                    & \textbf{5.6}            & \textbf{40.4}           & \textbf{0.758}       & $3.4$                     & $12.7$                    & $0.752$     
& $12.3$                    & $213.1$                   & $0.557$                                                                  \\

\hspace{7pt}HDBSCAN                            & $0.134$                          &                           & $11.7$                    & $216.5$                   & $0.738$                                    & $5.6$                     & $59.2$                    & $0.572$                                    & $10.6$                    & $195.0$                   & $0.520$                                                                       & $3.0$                     & $9.4$                     & $0.318$                     & $16.7$                    & $533.9$                   & $0.095$                \\
\hspace{7pt}k-means                                 & $0.481$                          &                           & \textbf{8.2}            & \textbf{107.2}          & \textbf{0.883}                           & $5.4$                     & $53.9$                    & $0.370$                                    & $9.1$                     & $140.4$                   & $0.735$            & $2.5$     & $8.9$  & $0.777$                              & $16.0$     & $415.3$    & $0.309$                                                                  \\
\hspace{7pt}OPTICS  & $0.406$ &    & $9.1$     & $148.4$ & $0.650$ & \textbf{3.8} & \textbf{21.2} & $0.704$   & $10.4$                    & $166.6$ & $0.458$   & $4.1$ & $20.2$ & $-4.564$    & $10.7$ & $184.4$ & $0.410$                               \\
\cline{2-18} 
\end{tabular}%
}
\caption{Prediction of questionnaire scores at $t_1$ given age, gender, score at $t_0$ (Baseline, first row of values) and, additionally the phenotype IDs returned by \COBALT{} (upper part), respectively by the clustering algorithms (lower part); for convenience, we ordered the columns with the same order as layers were added by \COBALT{}, but this ordering has no effect on the way the data are read by the predictor.
%AHC (agglomerative hierarchical clustering), BIRCH, GMM\_EM (Gaussian Mixture Models with Expectation Maximization algorithm), HDBSCAN, KMeans, OPTICS
}
\label{tab:predictions_best}
\end{table*}

\subsubsection{Prediction with the phenotypes of each \COBALT{} iteration}
The upper part of Table \ref{tab:predictions_best} shows for each iteration of \COBALT{} (first column) the quality of the phenotypes in the MLN network, measured as modularity Q (third column).
%We do not show the cost of adding a layer, since it was already observed that it increases at each iteration (cf. Figure \ref{fig:mod_layer_cost}). 
%
For each predicted score, we mark in \underline{underline} the best values achieved in a \COBALT{} iteration.
%Values that are also in \textbf{boldface} are the best throughout for a score (i.e. after the comparison with the clustering algorithms). Since the 1st iteration delivered the communities with the best modularity, we mark in \emph{italics} all cases where this iteration was second-best.

For each score, the \COBALT{}-augmented regressor outperforms the corresponding Baseline regressor, as can be seen by the \underline{underlined} values for the scores.
When focusing on the phenotypes of each iteration, we observe the following:
\begin{itemize}
    \item \textbf{iteration 1:}
    \COBALT{} outperforms the Baseline for the THI score and for the MDI and TBF12 scores (the MAE values are identical). For the TQ and TFI scores, the Baseline is better. 
    %Hence, the phenotypes of the THI layer alone suffice for predicting the scores of THI, MDI and TBF12 at $t_1$, with a better quality than a Baseline that has seen the scores of all 5 questionnaires at $t_0$.
    \item \textbf{iteration 2:}
    \COBALT{} outperforms the Baseline for the TQ, TFI and TBF12 scores. For the THI score, \COBALT{} is better with respect to MSE and MAE. 
    %Hence, the regressors using the \COBALT{} phenotypes on the THI and MDI layers suffice to predict 4 out of the 5 scores at $t_1$ with better MAE and MSE than the Baseline. \\
    For the MDI score, \COBALT{} is inferior than the Baseline. This is remarkable, since the layer included in the 2nd iteration is the MDI layer itself. However, as can be seen in Figure \ref{fig:iteration_2}, the communities are more oriented towards THI.
    \item \textbf{iteration 3:}
    \COBALT{} outperforms the Baseline for the TQ and TBF12 scores, and for the THI score with respect to MSE and MAE. With respect to the MDI score and the TFI score, the Baseline is superior.
    \item \textbf{iteration 4:}
    \COBALT{} outperforms the Baseline for the THI, MDI, TQ and TBF12 scores, i.e. all but the TFI score.
    \item \textbf{iteration 5:}
    \COBALT{} outperforms the Baseline for the THI score, but it is inferior to it for all the other scores.
\end{itemize}

Summarizing, the \COBALT{} phenotypes on the THI and MDI layers suffice to predict 4 out of the 5 scores at $t_1$ with better MAE and MSE than the Baseline that exploits the scores of all 5 questionnaires at $t_0$. The phenotypes of the THI layer alone suffice for predicting 3 out of 5 scores at $t_1$. This indicates that the exploitation of phenotypes during prediction is advantageous, and the advantage is higher when adding the least-cost layer.

The influence of the MDI layer must be perceived as an artifact, since this layer improves predictive performance but not for the MDI score itself An explanation is in the density of this layer, which may have resulted in poor-quality communities inside this layer.

\subsubsection{\COBALT{} vs clustering}
The lower part of Table \ref{tab:predictions_best} depicts the prediction quality achieved when using clustering algorithms for phenotype construction instead of \COBALT{} on MLNs. We varied the number of clusters to optimize silhouette, and we report the clusters found for this optimal number. For example, for k-means, the best silhouette was for $k=2$.

For each of the 5 scores there is at least one clustering algorithm that outperforms the Baseline (first row). As with \COBALT{}, this indicates that exploiting phenotypes during prediction is of advantage. However, unlike \COBALT{}, there is no clear winner among the clustering algorithms: for the THI score, all algorithms delivered models that were superior to the Baseline; for the TFI score, none did; for the other scores, some models were superior to the Baseline and others were inferior to it.

\COBALT{} outperforms the Clustering approaches when predicting the TFI score (iteration 2) and the TBF12 score (iteration 3), i.e. before the corresponding layers are added.
%\emph{ML} outperforms all clustering algorithms when predicting MDI score, TFI score and TBF12 score.
For the MDI score, iteration 4 delivers
%MLN with 4 layers provides 
the best $R^2$ value, but is inferior to OPTICS with respect to MSE and MAE.
%results; for the TFI score it is the 2-layer network and for the TBF12 score it is the 3-layer network. 
For the THI score, K-Means returns the best results. However, there is no clear winner among the Clustering approaches with respect to phenotype contribution: for each of the 5 scores, another algorithm is best for one or more of the three measures, although, unlike \COBALT{}, all Clustering algorithms are trained on all scores at $t_0$. In contrast, the phenotypes returned by \COBALT{} on the first two layers outperform the Baseline for all scores except the MDI.

\COBALT{} and Clustering cannot be compared directly on community homogeneity: for clustering, we use silhouette instead of modularity. When juxtaposing the results of the Clustering algorithms, we see that 
%For both types of algorithms the partition quality metric used (silhouette for the clustering algorithms and modularity for the community detection algorithm) do not agree with the evaluation performed with the regressors. I.e., iteration 1 does not achieve the best results, but it is the partition with the highest modularity. The same conclusion can be drawn from silhouette metric - 
BIRCH returns the phenotypes with
%is the algorithm with 
the highest silhouette value, but
%it outperforms the other ones only for the MDI score.
these phenotypes contribute less to prediction than those output by other clustering algorithms. This agrees with our observation on modularity for \COBALT{}: the phenotypes returned at iteration 1 have the highest modularity but the regressors exploiting them are of inferior quality.

Summarizing, the phenotypes returned by \COBALT{} exhibit more consistent predictive performance than achieved by clusters built by individual clustering algorithms, albeit the latter exploit all scores at $t_0$. This holds particularly for iteration 2, where a layer is chosen in a cost-sensitive way, before the stopping criterion SC1 is triggered. Phenotype quality, either the modularity in MLNs or the silhouettes in clustering, are poor indicators of predictive performance. For \COBALT{}, phenotypes built in a cost-sensitivity way lead to competitive predictive performance.

\subsubsection{\COBALT{} vs cost-insensitive MLN phenotypes}
In our earlier work \cite{Puga2021}, we built MLN-based phenotypes, optimizing on modularity rather than cost. Table \ref{tab:dsaa_results}, from \cite{Puga2021}, shows the layer selected at each iteration and the modularity scores achieved. 

\begin{table}[htb]
\begin{tabular}{lll}
\cline{2-3}
                     & \textbf{Layers}                & \textbf{Q} \\ \cline{2-3} 
\textrm{}{iteration 1} & THI                            & 0.752      \\
\textrm{iteration 2} & {[}THI, TQ{]}                  & 0358       \\
\textrm{iteration 3} & {[}THI, TQ, TBF12{]}           & 0.362      \\
\textrm{iteration 4} & {[}THI, TQ, TBF12, TFI{]}      & 0.002      \\
\textrm{iteration 5} & {[}THI, TQ, TBF12, TFI, MDI{]} & 0.001      \\ \cline{2-3} 
\end{tabular}
\caption{From \cite{Puga2021}: Modularity per iteration.}
\label{tab:dsaa_results}
\end{table}

When we compare with the column on $Q$ of Table \ref{tab:predictions_best}, we see
%To compare with to our previous approach \cite{Puga2021}, we show Table \ref{tab:dsaa_results} with the modularity values of each iteration. We observe
%%may infer 
that \COBALT{} achieves higher modularity values. %\textcolor{MyraFGcolor}{and better values for MSE, MAE and $R^2$ when predicting the 5 scores.}
This can be explained by the differences in the data:
%The differences can be accounted as follows: 
albeit we use the same dataset and prediction tasks, in \cite{Puga2021} we considered 
%However, there are two major differences between both approaches: the search algorithm (which is enhanced in the current work) and the data used. In \cite{Puga2021}, the original dataset was the same but 
only patients that had available data in all layers, whereas \COBALT{} uses all data, allowing for missing values in some layers. %from the original dataset.
%%Regardless of a considerable improvement, the following factors may have contributed to it: 
%Hence, 
%(i) the fact that \COBALT{} handles missing nodes in certain layers and (ii) the cost-sensitive core which influences the decision on which layers to add next, lead to models of higher quality.

A comparison on prediction quality is not possible because in \cite{Puga2021} we predicted for each community separately, and only for the TQ score at $t_1$, considering only communities that had enough data for learning and testing. Nonetheless, we can state that \COBALT{} is superior to the predecessor approach of \cite{Puga2021} by design, since it allows for missing values.

%Concerning the differences in the decisions on which layers to add: in \cite{Puga2021}, the layer chosen in the second iteration was the TQ, but in the current work was the MDI. Furthermore, the modularity of the second iteration was $0.358$ (cf. Figure 7 in \cite{Puga2021}), whereas COBALT achieves a modularity of $0.632$ at the same iteration.

% subsection on the communities of the 1st iteration
%\input{1stIterationCommunities}

\subsection{Impact of missingness} % on \COBALT{}}
\begin{figure}[htb]
\centering
       \includegraphics[height=8.2cm,keepaspectratio]{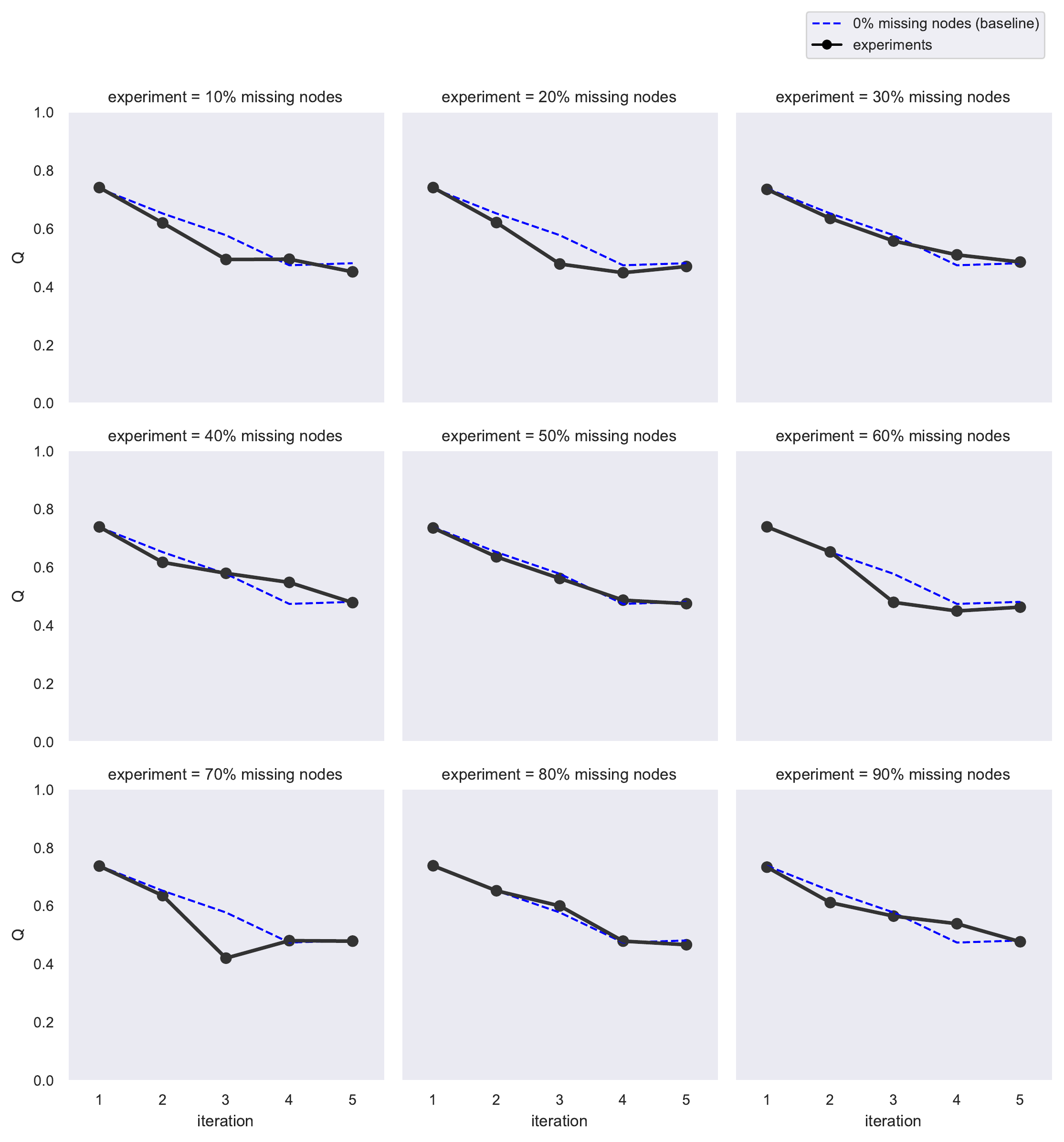}         
\caption{Modularity per iteration as missingness ratio is increased from 10\% to 90\%.
%and per missing node ratio
}
\label{fig:mv_ratio}
\end{figure}

%\subsection{The impact of missing nodes on the discovery of communities}
\label{missingness_results}
Figure \ref{fig:mv_ratio} shows how modularity changes from the 1st to the last iteration as we increase missingness from 10\% (leftmost, uppermost subfigure) to 90\% (rightmost, lowermost subfigure) in steps of percentual points.
%It is stressed that to increase missingness in a controlled way, we start with a much smaller subset of the data used for the previous analyses:
The data with 0\% missingness lead to the modularity values depicted by the dashed line in each of the subfigures.
%%the evolution of modularity (Q) over the iterations (each iteration corresponds to a layer added) and with different ratios of missing nodes from the original dataset.

\begin{table*}[tb]
\resizebox{\textwidth}{!}{%
\begin{tabular}{lllllllllllllllll}
\cmidrule(lr){2-17} \multicolumn{1}{c}{} & \multicolumn{1}{c}{\textbf{Partition quality}} & \multicolumn{5}{c}{\textbf{THI score at $t_1$}}                                                                                                           & \multicolumn{5}{c}{\textbf{MDI score at $t_1$}}                                                                                                           & \multicolumn{5}{c}{\textbf{TQ score at $t_1$}}                                                                                                            \\  \cmidrule(lr){2-2} \cmidrule(lr){3-7} \cmidrule(lr){8-12}  \cmidrule(lr){13-17} 
\multicolumn{1}{c}{}                                                        
& \multicolumn{1}{c}{Q (best)} & n  & NL & \multicolumn{1}{c}{MSE} & \multicolumn{1}{c}{MAE} & \multicolumn{1}{c}{$R^2$} & n & NL& \multicolumn{1}{c}{MSE} & \multicolumn{1}{c}{MAE} & \multicolumn{1}{c}{$R^2$} & n  & NL & \multicolumn{1}{c}{MSE} & \multicolumn{1}{c}{MAE} & \multicolumn{1}{c}{$R^2$} \\ \cmidrule(lr){2-2} \cmidrule(lr){3-7} \cmidrule(lr){8-12}  \cmidrule(lr){13-17} 
\multicolumn{1}{l}{Ratio of node missingness}  \\     

\hspace{7pt} $0\%$                                                                                   & $0.740$                                          & $48$ & $1$ & $8.0$   & $102.3$    & $0.850$        & $47$   &                   $1$                 &           $7.1$              &                   $90.0$   &                  $0.601$                           & $29$   &                                                     $1$ & $7.5$  & $88.6$ & $0.703$   \\
\hspace{7pt} $10\%$                                                                                   & $0.741$                                   & $48$ & $4$                                                       & $9.1$                     & $149.0$                   & $0.789$                                    & $47$ & $1$                                                       & $5.5$                     & $45.4$                    & $0.574$                                    & $29$ & $1$                                                       & $9.8$                     & $142.5$                   & $0.635$                                    \\
\hspace{7pt} $20\%$                                                                                   & $0.742$                                          & $48$ & $4$                                                       & $7.9$                     & $89.9$                    & $0.847$                                    &    &                                                         &                         &                         &                                          &    &                                                         &                         &                         &      \\
\cline{2-17}

\end{tabular}%
}
\caption{Prediction quality for each percentage value of missingness, showing the iteration (equiv. number of layers NL) that achieves the best modularity. 
%Prediction of questionnaire scores given age, gender and score at admission for the missing node ratios with a feasible number of training points. N: number of training data points, NL: number of layers.
}
\label{tab:prediction_mv}
\end{table*}

The curves indicate that the modularity is not greatly affected by missingness: it always drops towards 0.4, it always remains close to the reference (dashed) line of 0\% missingness, mostly a bit below it but sometimes above it. 
%The modularity values at iteration 1 assume close values for all missing node ratios. It is worth noting that there are some iterations in the experiments where the modularity in datasets with missing nodes is higher than in the baseline situation. The inverse also occurs.
These findings suggest that subsets of the nodes in the MLN have 
%properties (in terms of edge weights and edge existence) 
intra/inter-layer edges and weights that lead to communities of comparable quality to that of 0\% missingness. Table \ref{tab:prediction_mv} depicts prediction quality as we increase missingness from 0\% to 10\% and then to 20\%. Prediction was possible only for THI, MDI and TQ scores at $t_1$ when the missingness was 0\% and 10\%, and only for THI when the missingness was increased to 20\%. For TFI and TBF12, there were not enough data for training and testing. For each missingness value we chose the iteration with the highest modularity: this is reflected in the column `NL' that denotes the Number of Layers (NL) used and may differ depending on the questionnaire score being predicted. The number of patients used for training is shown in the column marked as `n', which again may differ for each questionnaire score.

In Table \ref{tab:predictions_best} we have seen that a decrease in modularity is not necessarily associated with a tendency in the values of the prediction quality measures. This agrees with the results in Table \ref{tab:prediction_mv}, where the sets of communities with the best modularity are chosen but the prediction quality measures vary in both directions. In general, the prediction quality is good, which indicates that a small increase in missingness does not lead to dramatic quality deterioration. However, the amount of training data is so small that a generalization is not possible.

\section{Conclusion}\label{conclusion}
In this work we have presented COBALT, a cost-based model to find communities in a single- or multi-layer network structure. We define cost as the cost of acquiring features and test it in a dataset with questionnaire data from chronic tinnitus patients. We also compare our results with traditional clustering methods.

The major findings of this work can be summarized into four:
\begin{enumerate}
    \item COBALT is able find partitions in the data that are superior to conventional clustering algorithms in terms of our evaluation criteria (performance on post-treatment data prediction) for three of the five questionnaire predictions;
    \item COBALT outperforms our prior work with the same dataset (cf. \cite{Puga2021}) by achieving higher modularity values. As a result, taking a cost-effective strategy and allowing missing values in each layer proved advantageous;
    \item The partitions with the best modularity do not lead necessarily to the best inputs for a post-treatment data predictor;
    \item Missing values have no substantial impact on the modularity of the partitions.
\end{enumerate}

% added paragraph on real world implementations
In the context of clinical decision support, \COBALT{} is intended for personalized diagnostics and treatment planing on the basis of patient phenotypes. \COBALT{} demonstrates that it is possible to build predictive phenotypes without demanding a large number of questionnaires, as was the case in our earlier work \cite{Niemann2020phenotypes}. Our approach can be used in different ways. First, we have shown that phenotypes are predictive; hence, once a patient's phenotype has been assessed, treatment outcome can be predicted without demanding that the patient fills further questionnaires. Thus, the burden of the patient during the diagnostic procedure can be reduced. Next, the acquisition of information during screening can be focused towards informative features, thus reducing cost without compromising quality. Finally, phenotype-specific treatments can be designed. To this purpose, clinical research is needed to assess the robustness of the identified models and their predictiveness. Moreover, the cost-sensitive selection of questionnaires and the usage of phenotypes in the clinical practice demand a modification of the diagnostic guidelines, which, again, is best initiated with a prospective study.

% Future work should concentrate on expanding the evaluation datasets and on the analysis of the impact of missing nodes on the discovery of communities. Another important track to investigate is the evaluation phase - different types of evaluation may be useful to complement the ones proposed.

\section*{Funding}
This project has received funding from the European Union's Horizon 2020 Research and Innovation Programme under grant agreement number 848261.

\section*{Conflict of interest statement}
The authors declare no conflict of interest.

%\begin{appendices}

%\section{Section title of first appendix}




%\end{appendices}

\bibliographystyle{plainnat} % make sure this is the right format
\bibliography{sn-bibliography}% common bib file
%\input sn-article.bbl

%\input sn-sample-bib.tex%

\end{document}